\documentclass[sigconf]{acmart}

\AtBeginDocument{%
  }

\copyrightyear{2023} 
\acmYear{2023} 
\setcopyright{acmlicensed}\acmConference[MM '23]{Proceedings of the 31st ACM International Conference on Multimedia}{October 29-November 3, 2023}{Ottawa, ON, Canada}
\acmBooktitle{Proceedings of the 31st ACM International Conference on Multimedia (MM '23), October 29-November 3, 2023, Ottawa, ON, Canada}
\acmPrice{15.00}
\acmDOI{10.1145/3581783.3612015}
\acmISBN{979-8-4007-0108-5/23/10}

\usepackage{balance}
\usepackage{multirow}
\usepackage[ruled,vlined]{algorithm2e}
\usepackage{bm}
\usepackage{fancyhdr}

\newcommand{\proposed}{{DWSF}}

\begin{document}

\title{Practical Deep Dispersed Watermarking with Synchronization and Fusion}

\author{Hengchang Guo}
\affiliation{%
  \institution{ByteDance Inc.}
  \city{Beijing}
  \country{China}}
\email{guohengchang@bytedance.com}
\author{Qilong Zhang}
\affiliation{%
  \institution{ByteDance Inc.}
  \city{Hangzhou}
  \country{China}}
\email{zhangqilong.ai@bytedance.com}
\author{Junwei Luo}
\affiliation{%
  \institution{ByteDance Inc.}
  \city{Beijing}
  \country{China}}
\email{luojunwei@bytedance.com}
\author{Feng Guo}
\authornote{Corresponding author.}
\affiliation{%
  \institution{ByteDance Inc.}
  \city{Beijing}
  \country{China}}
\email{guofeng.659@bytedance.com}
\author{Wenbin Zhang}
\affiliation{%
  \institution{ByteDance Inc.}
  \city{Hangzhou}
  \country{China}}
\email{zhangwenbin.hi@bytedance.com}
\author{Xiaodong Su}
\affiliation{%
  \institution{ByteDance Inc.}
  \city{Beijing}
  \country{China}}
\email{suxiaodong.sxd@bytedance.com}
\author{Minglei Li}
\affiliation{%
  \institution{ByteDance Inc.}
  \city{Beijing}
  \country{China}}
\email{liminglei@bytedance.com}

\renewcommand{\shortauthors}{Hengchang Guo et al.}

\begin{abstract}
Deep learning based blind watermarking works have gradually emerged and achieved impressive performance.
However, previous deep watermarking studies mainly focus on fixed low-resolution images while paying less attention to arbitrary resolution images, especially widespread high-resolution images nowadays. Moreover, most works usually demonstrate robustness against typical non-geometric attacks (\textit{e.g.}, JPEG compression) but ignore common geometric attacks (\textit{e.g.}, Rotate) and more challenging combined attacks.
To overcome the above limitations, we propose a practical deep \textbf{D}ispersed \textbf{W}atermarking with \textbf{S}ynchronization and \textbf{F}usion, called \textbf{\proposed}. 
Specifically, given an arbitrary-resolution cover image, we adopt a dispersed embedding scheme which sparsely and randomly selects several fixed small-size cover blocks to embed a consistent watermark message by a well-trained encoder.
In the extraction stage, we first design a watermark synchronization module to locate and rectify the encoded blocks in the noised watermarked image.
We then utilize a decoder to obtain messages embedded in these blocks, and propose a message fusion strategy based on similarity to make full use of the consistency among messages, thus determining a reliable message.
Extensive experiments conducted on different datasets convincingly demonstrate the effectiveness of our proposed {\proposed}. 
Compared with state-of-the-art approaches, our blind watermarking can achieve better performance: averagely improve the bit accuracy by 5.28\% and 5.93\% against single and combined attacks, respectively, and show less file size increment and better visual quality. 
Our code is available at https://github.com/bytedance/DWSF.
\end{abstract}


\maketitle

\section{Introduction}
Blind watermarking aims to embed specific identification information into multimedia contents (\textit{e.g.}, images, videos) in an invisible way, which is widely used for copyright protection and leak
source tracing at present. 
Moreover, in the context of the rapid development of Artificial Intelligence Generated Content (AIGC), blind watermarking can also be applied to identify AIGC, which can help to prevent the abuse of such generated contents. Therefore, blind watermarking has a very promising application prospect.

Following previous works~\cite{van1994digital,kang2003dwt,zhu2018hidden,fernandez2022watermarking}, we also focus on robust blind image watermarking to present our method.
In this case, blind watermark can be regarded as some kind of noise added to images, which is similar to adversarial perturbation~\cite{guo2021feature,wang2021admix,zhang2022practical,wang2022triangle,yuan2022natural,long22frequency}. Therefore, if the watermarked image is deliberately processed by image editing software during transmission, the watermark will inevitably be distorted.
A trivial solution to mitigate this issue is to increase the strength of the blind watermark, but this would sacrifice the visual quality of the watermarked image.
To find a way out of this dilemma, extensive research has been proposed in recent decades, which has also driven the booming development of robust blind watermarking.
Traditional approaches~\cite{van1994digital,kang2003dwt,barni1998adct,wan2022comprehensive,boussif2021novel,ko2020robust} typically embed watermark information in the original spatial domain or transform domain.
These methods rely heavily on hand-crafted features, thus generally having limited robustness against complex attacks~\cite{allwadhi2022comprehensive,wan2022comprehensive,mahto2021survey}. 
To alleviate this issue, various deep learning based bind watermarking works (\textit{e.g.}, HiDDeN~\cite{zhu2018hidden}, TSDL~\cite{liu2019novel}, MBRS~\cite{jia2021mbrs} and SSLW~\cite{fernandez2022watermarking}) have been proposed in recent years.
With the advantage of the end-to-end nature of deep learning and rich data augmentation, 
these approaches are capable of learning a more wise way to embed watermark information adaptively, thus achieving great robustness over traditional approaches.

\begin{figure*}
\centering
\setlength{\abovecaptionskip}{0.1cm} 
\includegraphics[width=0.98\linewidth]{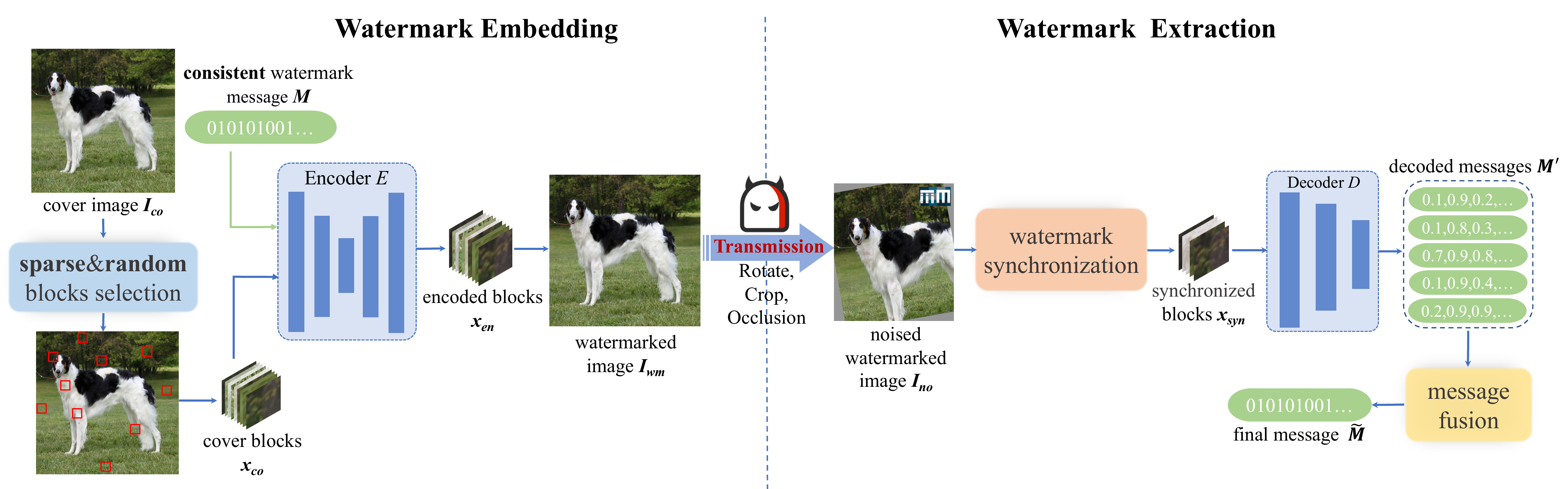}
\caption{The framework of our proposed {\proposed}. In the watermark embedding stage, a dispersed embedding scheme is adopted, \textit{i.e.}, several cover blocks are randomly and sparsely selected from the cover image and embedded with a consistent watermark message by a well-trained encoder.
In the transmission, the watermarked image may be attacked by various manipulations (\textit{e.g.}, Rotate, Crop, Occlusion) using image editing software. 
To facilitate message extraction from such a noised watermarked image, we design a watermark synchronization module to locate and rectify (\textit{i.e.}, inverse geometric transformation) the encoded blocks before decoding. 
Finally, the message fusion is proposed to determine a final message by analyzing multiple decoded results.
}
\label{fig:framework}
\end{figure*}

However, we find that two important aspects are not well studied in current deep learning based works, making them not practical enough in the real scenario. 
Firstly, most studies~\cite{jia2021mbrs,liu2019novel} present the evaluation on fixed low-resolution images, such as 128 $\times$ 128, which is obviously unreasonable. 
With the development of hardware and software technologies in recent decades, 
the resolution of photos has been growing rapidly, which is also reflected in the increase of image resolution in popular public datasets. Take OpenImages~\cite{alina2018open} as an example. Its image resolution varies from 320$\times$240 to 5000$\times$6000, and most images have a resolution greater than 2000 $\times$ 2000 (see Sec.~\ref{appendix:a} in Appendix). 
Therefore, it is worth taking account of arbitrary resolution (especially high resolution) images when applying deep learning based blind watermarking in the real scenario nowadays.
Secondly, the robustness of existing works is mainly tested against several typical non-geometric attacks (\textit{e.g.}, JPEG, Dropout), while paying less attention to common geometric attacks (\textit{e.g.}, Resize, Rotate, Padding). 
Different from non-geometric attacks that attenuate the strength of the watermark, geometric attacks destroy the watermark in a completely different way, \textit{i.e.}, leading to the desynchronization problem~\cite{osama2019attacking}.
More importantly, compared with single attacks, combined attacks (\textit{e.g.}, Rotate\&JPEG, Resize\&Crop) appear more frequently in the real scenario and bring greater challenges, yet are never discussed in previous studies~\cite{ahmadi2020redmark,zhu2018hidden,jia2021mbrs,liu2019novel,fernandez2022watermarking}.
Both of these make existing deep learning based works less practical, and thereby it is crucial to explore these missing issues to make the emerging deep watermarking more practical.

In this paper, we propose a practical deep blind watermarking framework called \textbf{D}ispersed \textbf{W}atermarking with \textbf{S}ynchronization and \textbf{F}usion (\textbf{\proposed}), and the overview is illustrated in Fig.~\ref{fig:framework}. 
In the embedding phase, we propose a novel scheme to dispersedly embed the watermark into multiple sub-regions of the cover image. To be specific, we sparsely and randomly select several small-size cover blocks from the cover image, and then embed a consistent watermark message into each block by a well-trained encoder that has been jointly trained with a decoder and a noise layer against non-geometric attacks.
Such an embedding design involves less modification compared to embedding the watermark message into the whole image like previous works ~\cite{zhu2018hidden,jia2021mbrs,liu2019novel,fernandez2022watermarking}, making the watermarked image have better visual quality and less file size increment. Moreover, this can also help to evade local erasure attacks (\textit{e.g.}, Crop, Occlusion) naturally. 
In the extraction phase, given a watermarked image, we design a watermark synchronization module (refer to Fig.~\ref{fig:psm}) which aims to locate and rectify the dispersed encoded blocks. Specifically, we adopt a segmentation model to capture imperceptible and even distorted watermark features so as to predict regions of encoded blocks. If the watermarked image is distorted by geometric attacks (\textit{e.g.}, Rotate), the segmentation results can help to estimate geometric transformation parameters (\textit{e.g.}, rotation angle) and thus reverse geometric transformation to obtain synchronized encoded blocks. 
After that, we can utilize the decoder to extract the message embedded in each synchronized encoded block. 
Finally, a message fusion strategy based on similarity is proposed to make full use of message consistency, which can circumvent the negative impacts of possible biases in decoded messages and thus determine a more reliable watermark message.

In summary, our main contributions are as follows:
\begin{enumerate}
    \item We point out that two important and practical aspects are not well addressed in existing deep learning based works, \textit{i.e.}, embedding in arbitrary resolution (especially high resolution) images, and robustness against complex attacks.
    \item To overcome these limitations, we propose a blind watermarking framework (called {\proposed}) which mainly consists of three novel components, \textit{i.e.}, dispersed embedding, watermark synchronization and message fusion. 
    \item We conduct extensive experiments on three different datasets (\textit{i.e.}, ImageNet~\cite{deng2009imagenet}, LabelMe~\cite{bryan2007labelme} and OpenImages~\cite{alina2018open}) against 11 kinds of single attack and 6 kinds of combined attack, and the results demonstrate the effectiveness and practicality of our proposed {\proposed}.
\end{enumerate}

\section{Related Work}
\subsection{Blind Image Watermarking}
Blind image watermarking technology has developed for a long time. 
In 1994, Schyndel et al.~\cite{van1994digital} first proposed embedding messages by manipulating image pixels (\textit{i.e.}, the Least Significant Bits) in the spatial domain. However, this method is not robust and can be easily detected by statistical measures~\cite{Dumitrescu03detcetion,Fridrich01Detecting,Fridrich02Practical}.
Moreover, researchers also attempted to exploit the frequency domain, and those proposed methods would first apply certain transforms (\textit{e.g.}, discrete Fourier transform (DFT)~\cite{kang2003dwt}, discrete cosine transform (DCT)~\cite{barni1998adct},  discrete wavelet transform (DWT)~\cite{daren2001dwt}) to the cover image and then embed the watermark message in the transform domains. Compared with spatial domain watermarking, these frequency domain ones usually achieved better robustness~\cite{fares2020robust,kishore2020novel}.

Recently, deep learning based methods are becoming increasingly popular due to the impressive performance of neural networks in feature extraction. 
Zhu et al.~\cite{zhu2018hidden} proposed an end-to-end solution which constructs a widely followed auto-encoder architecture: the encoder encodes the watermark message into the image and the decoder tries to extract the message embedded in the watermarked image.
Liu et al.~\cite{liu2019novel} introduced a two-stage training framework, which is composed of noise-free auto-encoder training and noise-aware decoder-only training, to resist non-differentiable distortions.
RedMark~\cite{ahmadi2020redmark} adopted a diffusion watermarking framework based on fully convolutional residual networks and achieved better performance in terms of imperceptibility and robustness. 
MBRS~\cite{jia2021mbrs} proposed a novel training scheme---randomly selects one from real JPEG, simulated JPEG and Identity (\textit{i.e.}, no attack) as the noise layer---to enhance robustness against JPEG compression.
Vukotic et al.~\cite{fernandez2022watermarking} used a pre-trained self-supervised model to obtain a transform-invariant latent space and embed watermark into the space, thus being more robust against a broad range of attacks.

\subsection{Watermark Synchronization}
The desynchronization problem, which is usually caused by geometric transforms (\textit{e.g.}, Rotate, Resize), can significantly affect the synchronization between watermark embedding and extracting, thus leading to the failure of decoding. 
To address this issue, various watermark synchronization solutions have been proposed. For example, Pereira et al.~\cite{pereira2000robust} proposed embedding an additional template into the image so as to estimate the transform parameters and then reverse these transforms.
Lin et al.~\cite{lin2001rotation} embedded the watermark message in a geometric transform-invariant domain by applying Fourier-Mellin transform.
However, these traditional methods are often tailored to specific distortions, which may have great limitations in application.
Therefore, recent works turn to training a neural network for watermark synchronization.
Tancik et al.~\cite{tancik2020stegastamp} fine-tuned a semantic segmentation model to locate the watermarked image on white background against printing attacks, which achieved satisfying segmentation performance since there have obvious bounds between the background and the image.
Luo et al.~\cite{luo2022leca} designed a model to predict the scale ratio and offset of the watermarked image before decoding, yet lacking of the discussion on other geometric attacks, \textit{e.g.}, Rotate, Padding.

\begin{table*}[h]
\caption{Description of common geometric and non-geometric attacks. For the visualization of attacked images, we present in Appendix (Sec.~\ref{appendix:b}). 
To ensure a fair comparison, 
we set the same random seed to control the intensity of the attack consistently across all methods.}
\resizebox{0.98\linewidth}{!}{\begin{tabular}{lll}
\toprule
Type & Attacks  &  Description\\ \midrule
\multirow{5}{*}{Geometric} &  Resize & Randomly scale the $H\times W$ watermarked image to $r_1H\times r_2W$, $r_1, r_2\in (0.5,2)$   \\
& Crop & Randomly crop a $c_1H\times c_2W$ region from the watermarked image, $c_1, c_2\in (0.7,1)$\\
& Rotate & Randomly rotate the watermarked image with angle $a$, $a \in (-30^{\circ},30^{\circ})$\\
& Padding & Randomly pad around the watermarked image with length $(p_{upper}, p_{bottom}, p_{left}, p_{right})$, $p_{upper}, p_{bottom}, p_{left}, p_{right} \in (0,100)$\\
& Picture In Picture (PIP) & Randomly put the $H\times W$ watermarked image onto a $e_1H\times e_2W$ clean image (\textit{i.e.}, without  watermark), $e_1, e_2\in (1,2)$\\
\hline
\multirow{6}{*}{Non-Geometric} & JPEG & Randomly compress the watermarked image with quality factor $q$, $q \in (50,100)$ \\
& GN &  Randomly add the gaussian noise to the watermarked image with variance $v$, $v \in [3,4,5,6,7,8,9,10]$\\
& GF & Randomly blur the watermarked image by the gaussian kernel with kernel size $k_s$, $k_s \in [3,5,7]$ \\
& Color & Randomly color jitter the watermarked image with factor $f$, $f \in (0.5,1.5)$\\
& Dropout & Randomly replace $p\%$ pixels of the watermarked image with pixels at the corresponding position of the cover image, $p \in (0,30)$\\
& Occlusion & Randomly put a $o_1H\times o_2W$ clean image (\textit{i.e.}, without watermark) onto the $H\times W$ watermarked image, $o_1, o_2\in (0.25,0.5)$\\
\bottomrule
\end{tabular}}
\label{table:Attack}
\end{table*}

\section{Method}\label{sec:method}

\subsection{Dispersed Embedding}
\label{sec:scheme}
Our method aims to embed watermark messages into images with arbitrary resolution while achieving better robustness and visual quality.
To accommodate such a challenging scenario, we propose a special dispersed embedding scheme.
Specifically, we randomly select a set of non-overlapping $h\times w$ (the default input size for our watermarking model) 
cover blocks $\bm{x_{co}}$ from the cover image $\bm{I_{co}}$ to embed a consistent watermark message. 
To reduce the modification of the image (\textit{i.e.}, sparse), 
we control the total area proportion of the selected cover blocks $\bm{x_{co}}$ to the cover image $\bm{I_{co}}$ is less than a small value Q (\%).
It is conceivable that this embedding scheme can be applied to arbitrary resolution images and innately hold better robustness against some local erasure attacks. Take Crop as an example. Even if some parts of the watermarked image $\bm{I_{wm}}$ are cropped, the remaining part may still have several complete encoded blocks that can be used to decode messages. Furthermore, this embedding scheme can even circumvent challenging collusion attacks~\cite{su2002Novel} (see Sec.~\ref{appendix:c} in Appendix for more details).

To sum up, our proposed dispersed embedding has three unique properties that do not appear in previous methods~\cite{zhu2018hidden,liu2019novel,jia2021mbrs,fernandez2022watermarking}:
\begin{itemize}
    \item \textit{Sparse}. Unlike previous works that embed a watermark message into the whole image by rule, \textit{sparse} image blocks with a small fixed size have less scope for modification, making the resulting watermarked image more human-imperceptible and having less impact on file size.
    \item \textit{Random}. Although embedding the watermark message in a fixed position in the image makes it easier for us to locate and extract, it also increases the security risk---being exploited by attackers to erase the watermark. To avoid this, we \textit{randomly} select embedded regions for each image.
    \item \textit{Consistent}. Since our encoded blocks are scattered in the image and the message embedded in each block is \textit{consistent}, we can reduce the bias of the final message by making full use of the similarity among all extracted results, thus determining a more reliable message.
\end{itemize}

\subsection{Watermarking Model Architecture}
\label{sec:watermarking}
\begin{figure}
\centering
\includegraphics[width=1\linewidth]{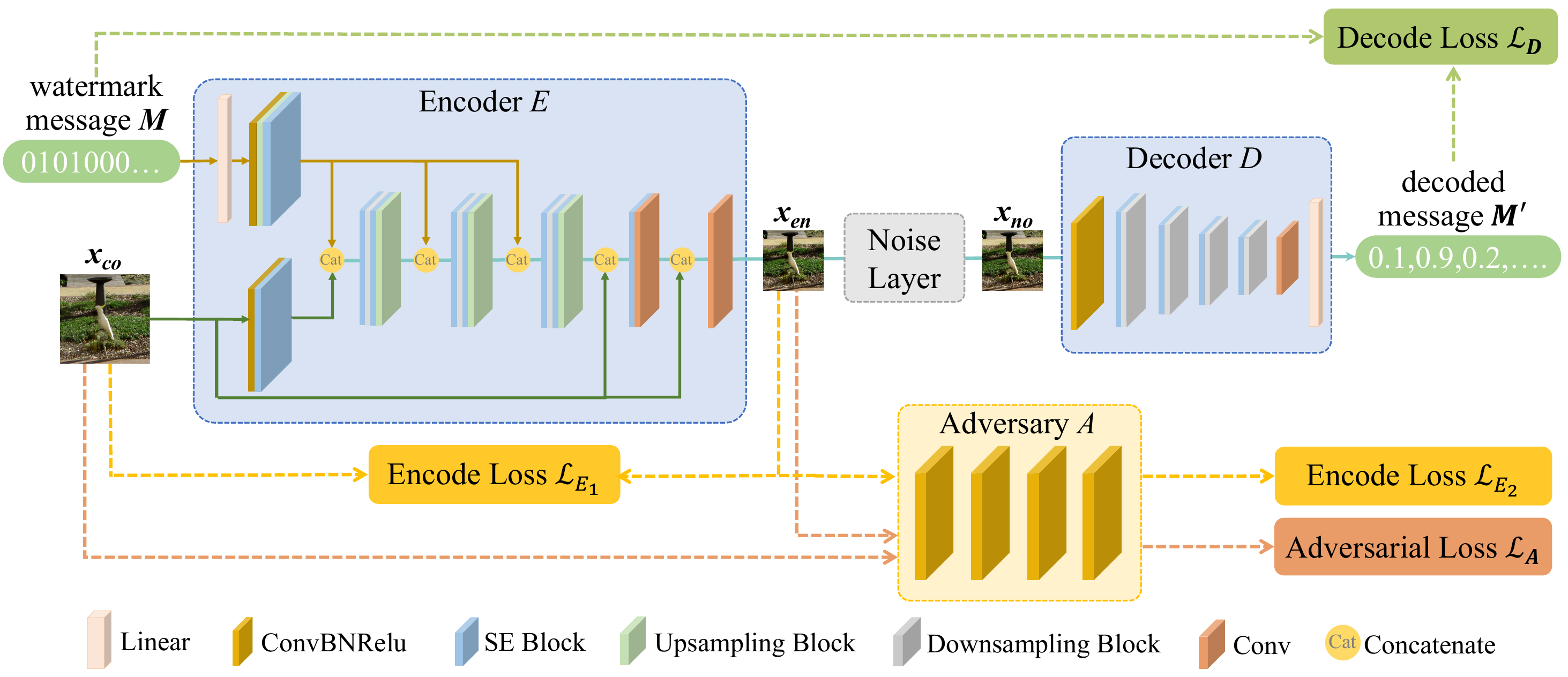}
\caption{Overall watermarking model architecture. The encoder embeds the watermark into the cover block. The noise layer mainly simulates non-geometric attacks. The decoder extracts the message from the noised block. The adversary distinguishes between the cover block and the encoded block.
}
\label{fig:encoder_decoder}
\end{figure}

Similar to previous works~\cite{zhu2018hidden,jia2021mbrs,liu2019novel}, we also train an end-to-end watermarking model to embed and extract watermark messages. 
The only difference is that the carrier of the watermark is not the whole image but the image block.
Our model architecture is illustrated in Fig.~\ref{fig:encoder_decoder}.
Formally, the encoder embeds the watermark message $\bm{M}$ into the cover block $\bm{x_{co}}$; the noise layer simulates common non-geometric attacks (\textit{e.g.}, JPEG, Dropout); the decoder learns to extract watermark message $\bm{M'}$ from the encoded block $\bm{x_{en}}$ or the noised encoded block $\bm{x_{no}}$; and the adversary discriminator aims to distinguish between $\bm{x_{co}}$ and $\bm{x_{en}}$, which can force the encoder to craft human-imperceptible watermarks.
The following part gives a detailed description of our watermarking model.

\textbf{Encoder.} 
The encoder takes a cover block $\bm{x_{co}} \!\in\![-1,1]^{3\times h\times w}$ and a binary watermark message $\bm{M} \!\in\! \{0,1\}^{L}$ with length $L$ as input, and aims to generate an encoded block $\bm{x_{en}}$ that is visually similar to $\bm{x_{co}}$. Here we refer to the network architecture of~\cite{liu2019novel,jia2021mbrs} to build our encoder. Specifically, we utilize Squeeze-and-Excitation (SE) module~\cite{hu2020squeeze} as the basic component of our encoder to integrate the watermark message into the image features. 
Besides, we not only repeatedly apply the concatenation operation~\cite{liu2019novel} to fully embed message features in shallow layers (\textit{i.e.}, brown solid lines), but also concatenate the raw image features in later layers to make the output maintain more image details (\textit{i.e.}, green solid lines), hence providing a good trade-off between robustness and imperceptibility.

To constrain the difference between the encoded block $\bm{x_{en}}$ and the cover block $\bm{x_{co}}$, we utilize two metrics to train our encoder.
One is the Mean Squared Error (MSE) loss, which controls the pixel-wise modification between $\bm{x_{en}}$ and $\bm{x_{co}}$. Another is the Multi-scale Structural Similarity for Image Quality (MSSSIM)~\cite{wang2003multiscale} which restricts the structure-wise changes between $\bm{x_{en}}$ and $\bm{x_{co}}$ at different scales. The overall loss function is shown as follows:
$$
\mathcal{L}_{E_1} =  MSE(\bm{x_{co}}, E_\theta(\bm{x_{co}},\bm{M})) +  \alpha * MSSSIM(\bm{x_{co}}, E_\theta(\bm{x_{co}},\bm{M})),
$$
where $\theta$ is the parameter of encoder $E$, $\alpha$ is the weight to balance MSE loss and MSSSIM loss.

\textbf{Noise Layer.} 
To address the challenge of watermark distortion in the real scenario, we deploy a noise layer as the data augmentation in the training stage to enhance the robustness of our watermarking model. 
Here, our noise layer mainly involves five non-geometric attacks (see Tab.~\ref{table:Attack}) and a noise-free identity mapping (called Identity). 
Since real JPEG is non-differentiable, we randomly choose one from JPEG-Mask~\cite{shin2017jpeg}, JPEG-SS~\cite{zhu2018hidden} and real JPEG instead in each iteration, which has been demonstrated to be effective in ~\cite{jia2021mbrs}.
In addition, unlike previous works~\cite{zhu2018hidden,liu2019novel,jia2021mbrs} that only consider attacks with a fixed strength (\textit{e.g.}, JPEG with quality=50), we suggest applying these attacks with random strength in the training stage. 
Such a setting can well deal with non-geometric attacks with unpredictable strength in the real scenario, making our method more practical.
For geometric attacks, we leave them to be handled by the watermark synchronization module proposed in Sec.~\ref{sec:location}.

\begin{figure*}[h]
    \centering
    \includegraphics[width=0.95\linewidth]{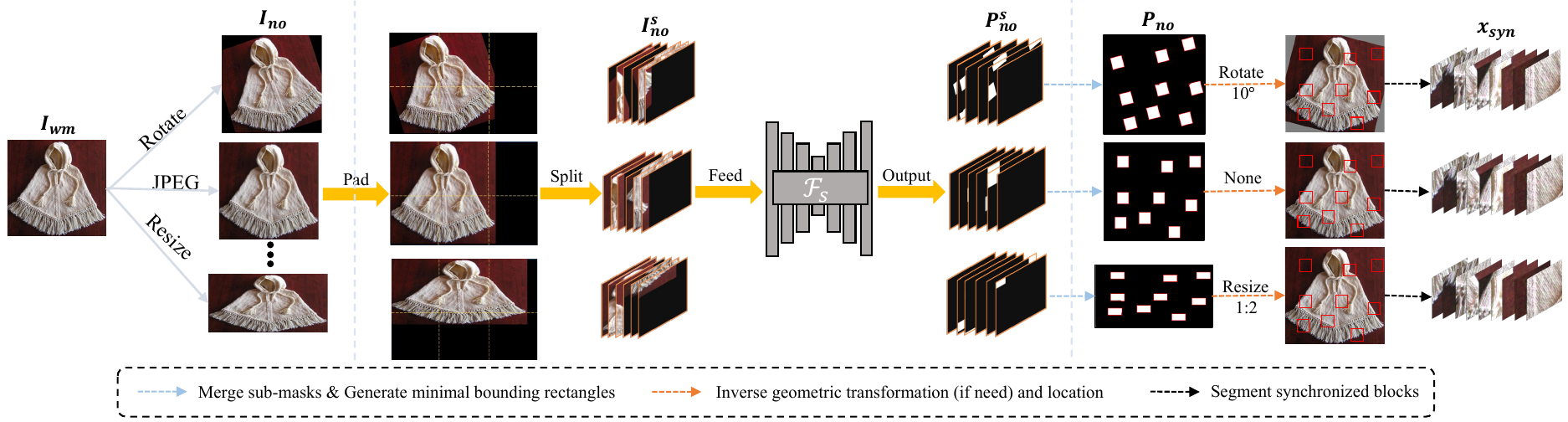}
    \caption{The pipeline for watermark synchronization module. Given a distorted watermarked image $\bm{I_{no}}$, we first utilize a watermark segmentation model $\mathcal{F}_s$ to locate encoded blocks in each sub-images $\bm{I_{no}^s}$, and then merge all sub-masks to yield a complete mask $\bm{P_{no}}$. By utilizing minimal bounding rectangles (\textit{i.e.}, red boxes) for predicted regions in $\bm{P_{no}}$, we can rectify deformations that may exist in $\bm{I_{no}}$, thus obtaining synchronized blocks $\bm{x_{syn}}$ for decoding messages.
    }
    \label{fig:psm}
\end{figure*}

\textbf{Decoder.} 
Given a noised encoded block $\bm{x_{no}}$ simulated via the noise layer, the task of the decoder is to extract the embedded watermark message in it. Here we also use SE modules to extract watermark features, and finally apply a linear layer to make the output have the same length (\textit{i.e.}, $L$) as the watermark message.
The objective of the decoder $D$ is to minimize the difference between the decoded message $\bm{M}'=D_\phi(\bm{x_{no}})$ and the original message $\bm{M}$, and the loss function is:
$$
\mathcal{L}_{D} = MSE(\bm{M}, D_\phi(\bm{x_{no}})),
$$
where $\phi$ is the parameter of decoder $D$.

\textbf{Adversary.} The task of the adversary discriminator is to further decrease the visibility of the watermark crafted via the encoder. 
In this work, we choose PatchGAN~\cite{phillip2017image} as the discriminator since it can encourage the encoded block to preserve more image details of the cover block.
To distinguish between the cover block and the encoded block, the adversary discriminator is optimized by minimizing the following loss:
$$
\mathcal{L}_{A} = log(1-A_\psi(E_\theta(\bm{x_{co}},\bm{M})))) + log(A_\psi(\bm{x_{co}})),
$$
where $\psi$ is the parameter of the adversary discriminator $A$. To evade the discriminator, the encoder should be updated by the following loss function:
$$
\mathcal{L}_{E_2} = log(A_\psi(E_\theta(\bm{x_{co}},\bm{M}))).
$$
Through such adversarial training, the encoder tends to craft indistinguishable encoded blocks, thus achieving better visual quality.

To sum up, the optimization objective for the encoder and decoder is to minimize $
\mathcal{L} = \lambda_{E_1}\mathcal{L}_{E_1} + \lambda_{E_2}\mathcal{L}_{E_2} +\mathcal{L}_{D}
$ ($\lambda_{E_1}$ and $\lambda_{E_2}$ are the weights to balance $\mathcal{L}_{E_1}$ and $\mathcal{L}_{E_2}$), and for the adversary is to minimize $\mathcal{L}_A$.

\subsection{Watermark Synchronization Module}
\label{sec:location}
As mentioned in Sec.~\ref{sec:scheme} and~\ref{sec:watermarking}, our decoder is designed for the fixed size (\textit{i.e.}, $h\times w$) encoded block, however, the watermarked image is arbitrary-resolution and embedded with multiple dispersed encoded blocks. Moreover, the watermarked image may be under various geometric attacks in transmission, which leads to the serious desynchronization problem.
Therefore, we propose a novel watermark synchronization module to locate and rectify these encoded blocks in the watermarked image before feeding them to the decoder. 
In the following, we will describe the pipeline of the watermark synchronization module in detail.

For the purpose of invisible watermarks (\textit{i.e.}, Sec.~\ref{sec:watermarking}), the visual difference between embedded and unembedded regions in the watermarked image is small. 
Nevertheless, the correlation among neighboring pixels in the embedded regions is inevitably destroyed, which can be captured by DNNs \cite{Boroumand2019Deep,Tan2021CALPA,castillo2021comprehensive}.
Inspired by it, we adopt the U$^2$-Net$^\dagger$~\cite{qin2020u2net}, which is lightweight but does well at capturing rich local and global information, to help segment encoded blocks and even noised encoded blocks.
To further improve performance, we consider not only the widely used Binary Cross-Entropy (BCE) loss but also the Intersection over Union (IoU) loss~\cite{qin2019basnet,qin2020u2net}. 
Thus, the overall loss of the segmentation model is formulated as:
$$\mathcal{L}_S=BCE(\bm{P^s},\bm{G^s})+\gamma IoU(\bm{P^s},\bm{G^s}), $$
where $\bm{P^s}\in [0,1]^{H_s\times W_s}$ is the predicted mask for the watermarked image, $\bm{G_{en}^s}\in \{0,1\}^{H_s\times W_s}$ is the ground truth of embedding mask, and $\gamma$ is a weight to balance two losses. 
In the real scenario, our watermark segmentation model should work for various resolution images---even if the ratio between the height and width of the image is arbitrary. Intuitively, it is challenging for the segmentation model because the gap between the fixed training image size and the arbitrary test image size is uncontrolled. A simple solution is to \textbf{scale} the test image to match the training size. However, this will raise a new gap: the additional scale operation induces additional distortion on the embedded regions, which may exceed the generalization capability of the segmentation model.
To tackle this issue, we propose a \textbf{Pad\&Split} strategy in the inference stage. The pipeline is illustrated in Fig.~\ref{fig:psm}.
Given a noised image $\bm{I_{no}} \in [0,1]^{3\times H\times W}$ that results from distorting the watermarked image $\bm{I_{wm}}$,
we first \textbf{pad} it to $(H+H_s-H\%H_s)\times (W+W_s-W\%W_s)$. Then, we \textbf{split} it to $\left\lceil\frac{H}{H_s}\right\rceil\cdot\left\lceil\frac{W}{W_s}\right\rceil$ sub-images so that the size of each sub-image is equal to the expected size ($H_s\times W_s$) of the watermark segmentation model.
By feeding them instead of scaled $\bm{I_{no}}$ to the segmentation model, we can obtain more precise predicted sub-masks $\bm{P_{no}^s}$. Then we merge all $\bm{P_{no}^s}$ to get complete predicted mask $\bm{P_{no}}$ (have cropped the pad part) for $\bm{I_{no}}$. 

The red boxes in $\bm{P_{no}}$ are corresponding minimum bounding rectangles for predicted regions. As we can observe, if the noised image is distorted by geometric attacks, the corresponding predicted mask $\bm{P_{no}}$ will keep the same deformation.
Thus, the noised image can utilize this to recover from the unknown geometric attacks.
Specifically, we first estimate geometric transformation parameters (\textit{e.g.}, rotation angle, scale factor) for each box in $\bm{P_{no}}$. Note our estimation is based on the prior knowledge that raw encoded block is $128\!\times\!128$ and without rotation. 
Then we filter out outliers (\textit{i.e.}, significantly different from others) and average the remaining estimated results to reduce bias. With the resulting parameters, we can rectify $\bm{P_{no}}$ and $\bm{I_{no}}$ back to the original state.
Finally, by mapping rectified bounding rectangles to the corresponding rectified image, we can gain expected synchronized blocks $\bm{x_{syn}}$ for our decoder, thus achieving the goal of watermark synchronization.

\subsection{Message Fusion}
\label{sec:fusion}
Unlike previous works that only decode a single message, our dispersed embedding in Sec.~\ref{sec:scheme} yield multiple decoded messages. 
To get a final message, an intuitive way is to average all decoded messages. However, this manner ignores the fact that several decoded messages may be substantially different from the true one under high-intensity attacks, thus causing bias in the final result. 

To avoid this problem, we propose a message fusion strategy based on message similarity to determine the final message $\bm{\Tilde{M}}$. The detailed algorithm is shown in Appendix Alg.~\ref{alg:message_fusion}. Formally, given $N$ decoded results $\bm{M}'\!\in\! [0.0,1.0]^L$, we calculate the difference (\textit{i.e.}, the number of inconsistent bits) between any two decoded messages:
$$
\mathcal{D}_{i,j}=\sum (Binary(\bm{M_i'})-Binary(\bm{M_j'}))^2,
$$
where $Binary(\cdot)$ binarizes each bit of the message with a threshold of 0.5.
Then we partition $\mathcal{D} \in [0,L]^{N^2}$ into $N$ sets, \textit{i.e.}, $\bm{S}=\{\bm{S_0}, \bm{S_1}, ..., \bm{S_{N-1}}\}$ where $\bm{S_i}=\{\mathcal{D}_{i,j}|j\in [0,N-1]\}$.
To prevent outlier messages from interfering with the final result, we set an upper limit of the bit difference threshold $T$. Specifically, we start $t$ from $0$ to $T$ to calculate the number of messages with bit difference less than $t$ in each set:
$$
\Tilde{i_t} = \mathop{\arg\max}\limits_{i\in [0,N-1]} |\bm{S_i} \leq t|,
$$
where $|\cdot|$ indicates the number of messages that satisfy the condition in each set $\bm{S_i}$. We stop the calculation once there exists a $t$ such that $|\bm{S_{\Tilde{i_t}}} \leq t| \geq K$, and the final message $\bm{\Tilde{M}}$ can be obtained as:
$$
\bm{\Tilde{M}}=Binary(Mean(\{\bm{M_j'}|\mathcal{D}_{\Tilde{i_t},j}\leq t\})),
$$
where $Mean(\cdot)$ averages each bit of the eligible messages.

\section{Experiments}
\subsection{Experimental Setting}
\,\,\,\,\,\,\,\textbf{Compared Methods.}
We compare our method with four state-of-the-art deep learning based watermarking methods (SOTAs), \textit{i.e.}, HiDDeN~\cite{zhu2018hidden}, TSDL~\cite{liu2019novel}, MBRS~\cite{jia2021mbrs} and SSLW~\cite{fernandez2022watermarking}.

\textbf{Dataset.}
Our training dataset is constructed by randomly sampling 40,000 images from the COCO dataset~\cite{lin2014coco}. Specifically, we randomly crop $128\!\times\!128$ (\textit{i.e.}, $h\!=\!w\!=\!128$) blocks in the images to train our watermarking model (\textit{i.e.}, the encoder, decoder and adversary) and SOTAs~\cite{zhu2018hidden,liu2019novel,jia2021mbrs}.
For our segmentation model, we randomly crop $512\!\times\!512$ (\textit{i.e.}, $H_s\!=\!H_w\!=\!512$) sub-regions in the watermarked images as its training dataset. To demonstrate the generalization of DWSF, we randomly sample 1,000 images with default resolution (\textit{i.e.}, not pre-scaled to a fixed size) from ImageNet~\cite{deng2009imagenet}, LabelMe~\cite{bryan2007labelme} and OpenImages~\cite{alina2018open} respectively as our testing data. Note that MBRS is an exception: it only accepts a fixed input size (here is $128\!\times\!128$) after finishing training. We need to scale the testing image to $128\!\times\!128$ when evaluating it.

\textbf{Metrics.}
In our paper, we consider peak signal-to-noise ratio (PSNR), Byte Increase Rate---the percentage of byte increase of watermarked compared to cover images, and Bit Accuracy to evaluate visual quality, file size increment and robustness, respectively.

\textbf{Implementation Details.}
Our method is implemented by PyTorch~\cite{paszke2019pytorch} and executed on a NVIDIA A100 GPU. We use the AdamW~\cite{adamw} optimizer with a learning rate of 1e-4, and set weight factor $\lambda_{E_1}$, $\lambda_{E_2}$, $\alpha$ and $\gamma$ to 0.2, 0.001, 0.005 and 0.1, respectively. Both watermarking model (with batch size 64) and segmentation model (with batch size 24) are trained for 100 epochs, while the compared methods adopt their default settings. The length $L$ of message $\bm{M}$ is 30. 
For dispersed embedding, the area proportion Q is 25\% but with an upper limit of 20 blocks if the image resolution is too high (discussion about this can be found in Appendix Sec.~\ref{appendix:d}).
For message fusion, the bit difference threshold $T$ is 5 and the smallest number $K$ is 2.
For the compared methods and our segmentation model, we consider both non-geometric and geometric attacks introduced in Tab.~\ref{table:Attack} to serve as the noise layer so that the model can learn to defend against them.
For our watermarking model, we mainly consider non-geometric attacks (JPEG, GN, GF, Color and Dropout, and Identity). We also deploy several geometric attacks with low strength to compensate for the error in segmentation.
For fairness, all methods clip the PSNR of their embedded regions to 35dB in the inference phase. Note the result of PSNR would fluctuate slightly due to the rounding operation (float$\rightarrow$unit8) when saving images.

\subsection{Comparison with SOTAs}
In this section, we comprehensively compare our {\proposed} with four state-of-the-art methods from three perspectives: \textit{visual quality}, \textit{file size increment} and \textit{robustness}. 
To demonstrate the generalization of our {\proposed}, we consider ImageNet, LabelMe and OpenImages datasets to conduct our experiments.

\begin{figure}
    \centering
    \includegraphics[width=0.98\linewidth]{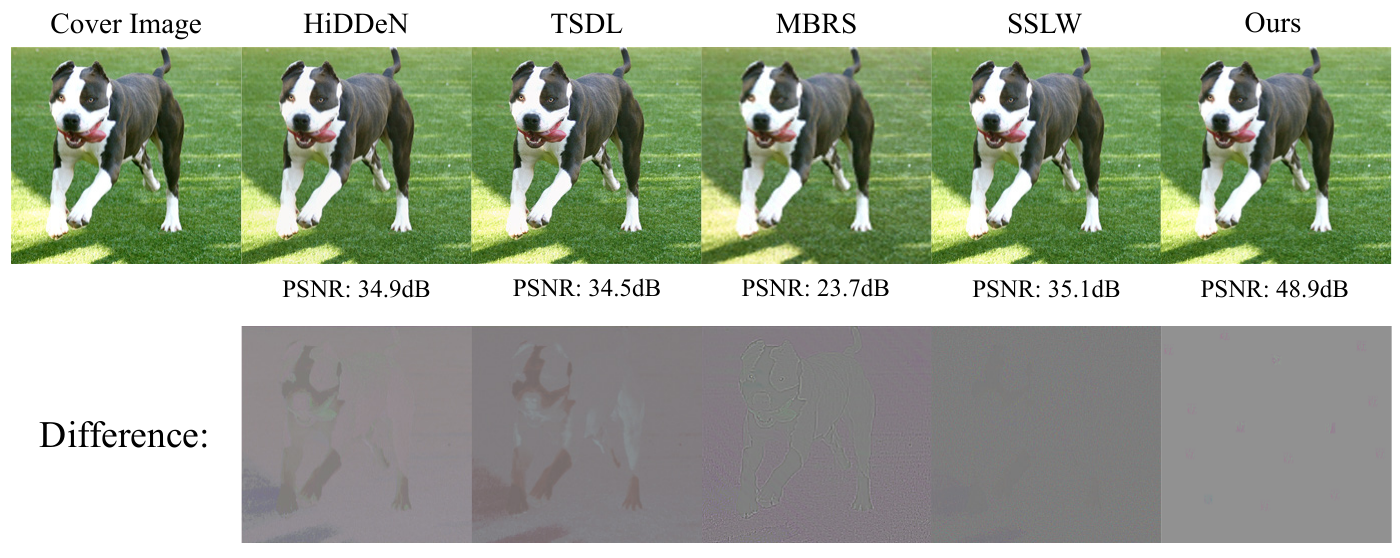}
    \caption{
    Visualization of watermarked images generated by different methods. The first row lists the watermarked images and the second row shows the difference between watermarked images and cover images. 
    }
    \label{fig:visual_quality}
\end{figure}

\begin{table}[!] 
\caption{Comparison on visual quality (PSNR) and file size increment (Byte Increase Rate).}
\resizebox{0.98\linewidth}{!}{\begin{tabular}{lcccccc}
\toprule
\multirow{3}{*}{Models} & \multicolumn{3}{c}{PSNR (dB) $\uparrow$}   & \multicolumn{3}{c}{Byte Increase Rate (\%) $\downarrow$}\\ \cmidrule(r){2-4} \cmidrule(r){5-7} 

& ImageNet  &  LabelMe  &  OpenImages    & ImageNet  &  LabelMe  &  OpenImages    \\
\midrule
HiDDeN & 35.90 & 35.22 & {35.58} & 10.27 & 12.17 & 29.08 \ \\
TSDL & 34.80 & 34.76 & 34.94 & {2.74} & {8.30} & 27.49 \\
MBRS & 24.14 & 23.96  & 24.14 & 9.20 & 11.95 & {8.82} \\
SSLW & 35.10 & 35.03 & 35.16 & 19.67 & 22.48 & 45.95\\
Ours & \textbf{42.07} & \textbf{47.38} & \textbf{46.12} &\textbf{2.41} & \textbf{1.30} &\textbf{1.90}\\
\bottomrule
\end{tabular}}
\label{table:psnr_and_volume_growth_rate}
\end{table}

\begin{table*}[h]
\caption{Comparison of Bit Accuracy (\%)  against single attacks.}
\resizebox{0.975\linewidth}{!}{
\begin{tabular}{clccccccccccccc}
\toprule
\multirow{2}{*}{Dataset} & \multirow{2}{*}{Models} & \multicolumn{12}{c}{Attacks} & \multirow{2}{*}{AVG.}                   \\ \cline{3-14} 
                  & & Identity & JPEG & GN & GF & Color & Dropout & Resize & Crop & Rotate & Padding & Occlusion & PIP        \\
\midrule
\multirow{5}{*}{ImageNet} 
& HiDDeN  & 95.19  & 72.72  & 93.47  & 80.96  & 94.83  & 94.46  & 90.19   & 95.02  & 92.39   & 94.65  & 94.78  & 92.45 & 90.93 \\
& TSDL & 99.26 & 52.46  & 97.82  & 55.44  & 98.37  & 98.11  & 75.33   & \textbf{99.27}  & 87.61   & 98.58  & 98.92  & 93.52 & 87.89 \\
& MBRS   & 97.97   & 97.16  & 97.96  & 97.67  & 97.74  & 97.11  & 97.96   & 68.13  & 68.68   & 68.38  & 95.71  & 57.39  & 86.82 \\
& SSLW  & \textbf{100.00}         & 96.60          & 99.92          & 99.42          & 99.87          & 62.11          & 84.89          & 98.33          & \textbf{98.49}          & 98.39          & \textbf{99.60}   & 81.63   &93.27       \\
& Ours   & \textbf{100.00}          & \textbf{98.11}          & \textbf{100.00}          & \textbf{99.78}          & \textbf{99.95}          & \textbf{99.96}          & \textbf{99.80}          & 98.61          & 98.20          & \textbf{99.99}          & 99.04   & \textbf{100.00}   &\textbf{99.45}      \\ 
\midrule
\multirow{5}{*}{LabelMe} &HiDDeN   & 97.19  & 73.21 & 96.26  & 84.35 & 97.08 & 96.94  & 93.16 & 97.19  & 95.84 & 97.16  & 97.10  & 94.15 & 93.30\\
& TSDL & 99.71 & 52.61 & 98.24  & 59.19 & 99.21 & 99.02 & 79.22 & 99.71 & 90.74 & 99.76 & 99.75 & 93.62 & 89.23\\
& MBRS     & 99.94  & \textbf{99.93} & 99.94  & 99.93 & 99.94 & 99.88  & 99.94 & 72.28  & 73.61 & 92.78  & 98.93  & 60.39 & 91.46\\
& SSLW      & \textbf{100.00} & 99.18 & \textbf{100.00} & 99.85 & 99.99 & 58.70  & 94.07 & 99.87  & 99.95 & 99.98  & 99.99  & 92.88 &95.37\\
& Ours     & \textbf{100.00}  & 99.74 & \textbf{100.00}  & \textbf{100.00} & \textbf{100.00} & \textbf{100.00}  & \textbf{100.00} & \textbf{100.00}  & \textbf{100.00} & \textbf{100.00}  & \textbf{100.00}  & \textbf{100.00} &\textbf{99.98} \\
\midrule
\multirow{5}{*}{OpenImages} & HiDDeN            & 96.71           & 71.70          & 94.86          & 82.91          & 96.46          & 96.18           & 93.24           & 96.70         & 95.14          & 96.62           & 96.54   & 92.34   &92.45        \\
                  & TSDL          & 99.52           & 52.65          & 97.84          & 58.98          & 98.80          & 98.83           & 78.62           & 99.60          & 90.18         & 99.27           & 99.46   & 88.69  &88.54         \\
                  & MBRS              & 97.58           & 97.45          & 97.55          & 97.54          & 97.30          & 96.78           & 97.55           & 67.15          & 68.12          & 81.72           & 95.45   & 57.74      &87.66     \\
                  & SSLW               & 99.99          & 97.96          & 99.82     & 99.70          & 99.70          & 61.05           & 93.82           & 99.69          & \textbf{99.70}          & 99.84           & \textbf{99.93}   & 87.16   &94.86        \\
                  & Ours              & \textbf{100.00} & \textbf{99.52} & \textbf{100.00} & \textbf{99.94} & \textbf{99.95} & \textbf{100.00} & \textbf{100.00} & \textbf{99.78} & 99.64 & \textbf{100.00} & 99.91   & \textbf{100.00} &\textbf{99.90}      \\  
\bottomrule
\end{tabular}
}
\label{table:BitACC_single}
\end{table*}

\begin{table*}[]
\caption{Comparison of Bit Accuracy (\%) against combined attacks.}
\resizebox{0.975\linewidth}{!}{
\begin{tabular}{lccccc|ccccc|ccccc}
\toprule
\multirow{3}{*}{Attacks} & \multicolumn{5}{c}{ImageNet}              & \multicolumn{5}{c}{LabelMe}   &     \multicolumn{5}{c}{OpenImages}                                  \\ \cmidrule(r){2-6} \cmidrule(r){7-11} \cmidrule(r){12-16}
& HiDDeN & TSDL & MBRS  & SSLW   & Ours   & HiDDeN & TSDL  & MBRS    & SSLW   & Ours   & HiDDeN & TSDL       & MBRS           & SSLW   & Ours          \\
\midrule
Color\&JPEG       & 72.19 &52.30 &\textbf{96.69} &94.40 &96.41     & 72.86&52.40&\textbf{99.87}&98.17&99.59          & 70.74 &52.67 &97.31 &96.70 &\textbf{98.87}     \\
Crop\&JPEG         & 70.76 & 52.15 & 67.65  & 88.74 & \textbf{95.47}        & 72.00  & 52.62  & 72.30  & 97.70 & \textbf{99.37}       & 70.08 & 52.65 & 67.63 & 95.16 & \textbf{98.88} \\
Crop\&Resize       & 89.57 & 75.27   & 67.64  & 81.09  & \textbf{96.72}   & 93.07 & 79.15  & 72.99  & 91.79 & \textbf{99.92}       & 92.29 & 79.09  & 62.27  & 90.52  & \textbf{99.86} \\
Occlusion\&JPEG      & 71.90  & 52.29   & 94.41 & 92.42  & \textbf{95.93}       & 71.64 & 52.32 & 98.84 & 98.41 & \textbf{99.64}       & 69.87  & 52.31   & 95.03   & 96.21 & \textbf{99.12} \\
Crop\&Resize\&JPEG  & 64.14 & 50.43  & 67.24   & 72.94 & \textbf{86.87}      & 65.44 & 50.47 & 72.14 & 86.04 & \textbf{93.06}       & 64.75 & 50.80  & 67.41  & 82.80 & \textbf{92.54} \\
Crop\&Occlusion\&JPEG & 69.37  & 52.04  & 66.04   & 82.38 & \textbf{90.75}    & 70.76       & 52.15   & 70.98   & 95.77 & \textbf{98.94}       & 68.26  & 52.55 & 65.97  & 92.14 & \textbf{98.28} \\
\midrule
AVG. & 72.99 & 55.75 &76.61 &85.33 &\textbf{93.69}  & 74.30 &56.52 &81.19 &94.65 &\textbf{98.42}  & 72.67 &56.68 &75.94 &92.26 &\textbf{97.93}  \\ 
\bottomrule
\end{tabular}
}
\label{table:BitACC_multiple}
\end{table*}

\subsubsection{Visual Quality \& File Size Increment}
Visual quality and file size increment are two important points for practical watermarking. Particularly, a high visual quality can hide the watermark to prevent it from being noticed, and a small file size increment can save transmission bandwidth and storage resources. To achieve the above goals, we propose a novel dispersed watermarking in Sec.~\ref{sec:scheme} which modifies only several sub-regions of the image instead of the whole image.
In this section, we use PSNR and Byte Increase Rate to quantitatively compare our method with SOTAs. For fairness, all images are saved as ``PNG" format.

The visual comparison is shown in Fig.~\ref{fig:visual_quality}, and the quantitative results are listed in Tab.~\ref{table:psnr_and_volume_growth_rate}. It can be observed that the resulting PSNR of HiDDeN, TSDL and SSLW are all around 35.0dB, while ours is much higher. For example, our method gets \textbf{47.38}dB on LabelMe dataset, which is 12.16dB higher than HiDDeN. Another observation from Tab.~\ref{table:psnr_and_volume_growth_rate} is that the PSNR of MBRS is very low.
This is because MBRS can only accept fixed input size after finishing training---we need first scale the cover image to the available size of MBRS (\textit{i.e.}, $128\times128$) and then scale the watermarked image back to its original size.
After these two scaling operations, details of the image will inevitably lose, especially when the original size of the image is much higher than the available size of MBRS.

Besides, as shown in Tab.~\ref{table:psnr_and_volume_growth_rate}, our {\proposed} has the minimal increment on image file size, with byte increase rate lower than \textbf{2.5\%}.
For other methods, the file size of encoded images increases significantly, especially when the resolution of original images is large.  For example, after embedding watermarks into OpenImages images via SSLW, the bytes of images will increase by over 45.0\% ($\bm{24\times}$ larger than ours), which causes much more storage space and higher transmission bandwidth.

\subsubsection{Robustness against Single Attacks}
In this section, we evaluate the robustness of our method and SOTAs against JPEG, GN, GF, Color, Dropout, Resize, Crop, Rotate, Padding, Overly, PIP (detailed description of these attacks is listed in Tab.~\ref{table:Attack}), and Identity.

Tab.~\ref{table:BitACC_single} reports the bit accuracy after attacks. Notably, our {\proposed} always achieves over \textbf{98\%} bit accuracy, and in most cases close to \textbf{100\%}, regardless of the dataset or the attack it faces. 
Although in some cases we cannot outperform the best method, we are still able to place $2^{nd}$ and not far behind the $1^{st}$. Take the result of OpenImages~\cite{alina2018open} as an example. When using Rotate to distort the watermarked image, {\proposed} can get 99.64\% bit accuracy, which is very close to the best one (\textit{i.e.}, 99.70\%) achieved by SSLW.
On average, {\proposed} outperforms HiDDeN, TSDL, MBRS and SSLW by \textbf{7.55\%}, \textbf{11.22\%}, \textbf{11.13\%} and \textbf{5.28\%} respectively, which convincingly indicates the generalization and robustness of our method.

\subsubsection{Robustness against Combined Attacks}
In the real scenario, watermarked images may be subject to more than one type of attack. 
Thus, the robustness of watermarking against combined attacks should also be investigated.
Since Color, Occlusion, Crop, Resize and JPEG are very common in the real scenario, we use them to simulate some combined attacks: Color\&JPEG, Crop\&JPEG, Crop\&Resize, Occlusion\&JPEG, Crop\&Resize\&JPEG, Crop\&Occlusion\&JPEG, and then evaluate the robustness of our {\proposed} and SOTAs.

As demonstrated in Tab.~\ref{table:BitACC_multiple}, the performance gap between the SOTAs and our proposed method is further enlarged. Among the SOTAs, only MBRS performs slightly better than our {\proposed} in two cases. Nonetheless, its visual quality is far worse than ours (see Fig.~\ref{fig:visual_quality}), which is impractical in the real scenario.
On average, our {\proposed} outperforms HiDDeN, TSDL, MBRS and SSLW by 
\textbf{23.36\%}, \textbf{40.36\%}, \textbf{18.77\%}, and \textbf{5.93\%}, respectively. This significant advantage is mainly attributed to our proposed \textit{dispersed embedding}, \textit{watermark synchronization} and \textit{message fusion}, which benefit the watermarking robustness in this challenging scenario.

\subsection{Ablation Study}
\subsubsection{Inference Strategy for Watermark Segmentation Model}
\label{sec:inference_strategy}
\begin{figure}
\centering
\includegraphics[width=1\linewidth]{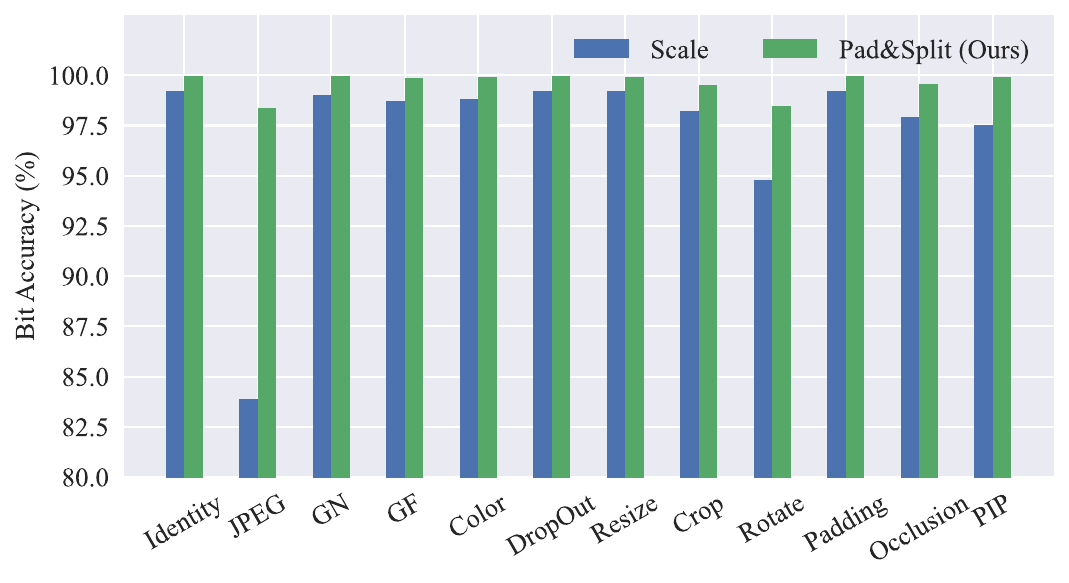}
\caption{Bit Accuracy (\%) of the Scale and our Pad\&Split strategies on ImageNet.}
\label{fig:infer_cmp}
\end{figure}

As mentioned in Sec.~\ref{sec:location}, we argue that scaling the test image to $H_s\times W_s$ beforehand is not a good solution for the segmentation model in the inference stage. To support this claim, we compare our proposed Pad\&Split preprocessing with the Scale preprocessing (the corresponding segmentation model is trained with scaled $H_s\times W_s$ images) on all attacks and Identity. 
The resulting bit accuracy is depicted in Fig.~\ref{fig:infer_cmp}. A first glance shows that our Pad\&Split consistently surpasses the Scale. Notably, when watermarked images are distorted by JPEG, our proposed Pad\&Split outperforms the Scale by \textbf{14.5\%}. This indicates that Pad\&Split is more effective, which can obtain more precise predicted masks for watermarked images, thus improving the bit accuracy of decoded results.

\begin{figure}
\centering
\includegraphics[width=1\linewidth]{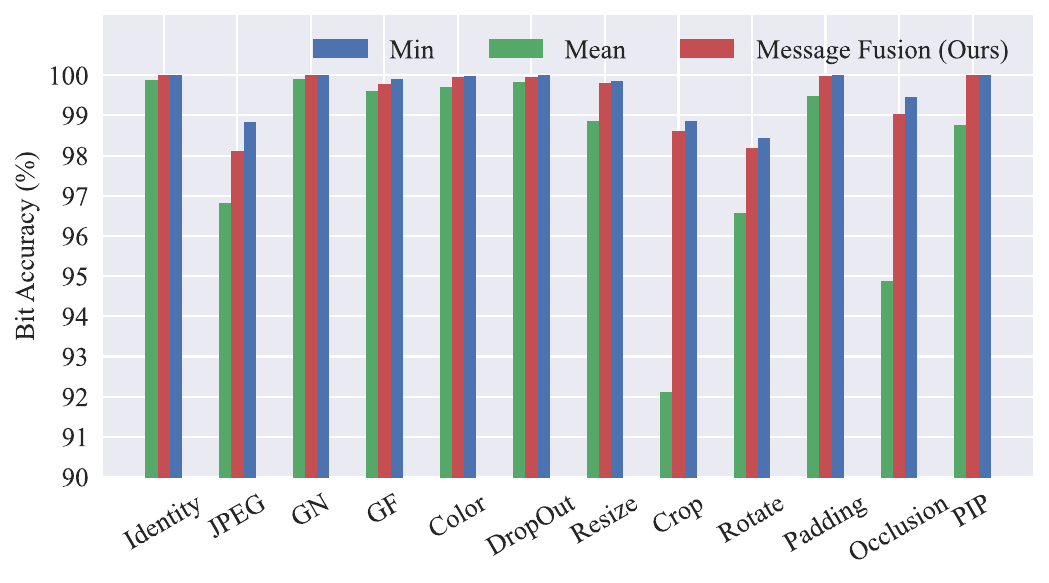}
\caption{Bit Accuracy (\%) of the Min, the Mean and our message fusion on ImageNet.
}
\label{fig:min}
\end{figure}

\subsubsection{Message Fusion}
In this paper, we propose a message fusion strategy to make full use of the similarity among multiple decoded messages. To demonstrate the effectiveness of this strategy, we compare it with Min and Mean strategies. Specifically, the Mean trivially
averages each bit of all the decoded messages, and the Min selects the best result from all the decoded messages to evaluate bit accuracy. Obviously, Min strategy is not practical in the real scenario since the true watermark message is unknown, but it reflects the theoretical upper bound. 

As depicted in Fig.~\ref{fig:min}, trivially adopting the Mean strategy is less effective than our message fusion. This is because our strategy can filter out results with low confidence (\textit{i.e.}, outlier) and thus reduce bias. On average, our message fusion strategy can achieve \textbf{99.45\%} bit accuracy, which is only 0.16\% lower than that of the Min strategy. This convincingly demonstrates that our message fusion strategy can achieve great performance close to the theoretical upper bound.

\begin{table}[]
\caption{Comparison of Bit Check Accuracy (\%) on ImageNet}
\resizebox{1\linewidth}{!}{
\begin{tabular}{lccccc}
\toprule
Attacks & HiDDeN & TSDL & MBRS  & SSLW   & Ours \\
\midrule
Identity & 40.70 & 85.30  & 86.70 & \textbf{100.00} & \textbf{100.00}\\
JPEG & 1.50 & 0.00 & 83.00 & 59.20 & \textbf{88.40}\\
Resize & 22.10 & 3.10 & 86.70 & 23.90 & \textbf{99.60}\\
Crop & 40.20 & 84.80 & 2.40 & 73.90 & \textbf{95.90} \\
Occlusion & 39.70 & 80.70 & 61.30 & 92.20 & \textbf{96.10}\\
Crop\&JPEG & 0.90 & 0.00 & 2.20 & 21.20 & \textbf{80.10}\\
Occlusion\&JPEG & 2.30 & 0.00 & 56.80 & 33.50 & \textbf{78.80} \\
Crop\&Occlusion\&JPEG & 0.90 & 0.00 & 1.90 & 8.00 & \textbf{64.60}\\
\bottomrule
\end{tabular}
}
\label{table:MessageACC_all}
\end{table}

\subsection{Discussion in the Real Scenario}
Bit accuracy is a widely used metric to measure the effectiveness of blind watermarking. Nonetheless, it is inappropriate in practice since the real scenario requires all bits in the message to be correct. To confirm the decoded message is valuable, we introduce a new metric called bit check accuracy by using 8-bits Cyclic Redundancy Check (CRC)~\cite{peterson1961cyclic}. Formally, bit check accuracy defines a decoded message is correct only when the CRC value can match, otherwise is wrong. Note that the previous methods only output a single message for a cover image, while ours has more ($N+1$ messages from $N$ synchronized blocks and $1$ result derived from message fusion) messages that can be checked---higher fault tolerance.

Tab.~\ref{table:MessageACC_all} gives the result of bit check accuracy against several attacks on ImageNet dataset.
Take the results under Resize attack in Tab.~\ref{table:BitACC_single} and Tab.~\ref{table:MessageACC_all} as an example. We can observe that the bit check accuracy of HiDDeN, TSDL, MBRS and SSLW is much lower than corresponding bit accuracy---significantly reduce from 90.19\%, 75.33\%, 97.96\%, 84.89\% to 22.10\%, 3.10\%, 86.70\%, 23.90\%, respectively. 
In fact, for a 30-bit message, even if the bit accuracy can reach 96.67\% (\textit{i.e.}, only one bit is wrong), the bit check accuracy is 0\%, still worthless in practice.
In contrast, the bit check accuracy of our method (\textit{i.e.}, 99.60\%) is very close to corresponding bit accuracy (\textit{i.e.}, 99.80\%), which highlights the potential of our approach in the real scenario.

\section{Conclusion}
In this paper, we propose a novel watermarking framework ({\proposed}) to enhance robustness against various attacks, in which an auto-encoder is trained to be robust against non-geometric attacks and a watermark synchronization module is designed to resist geometric attacks.
Meanwhile, a dispersed embedding scheme is proposed to make the approach be applied to arbitrary-resolution images, especially high-resolution images, and a message fusion strategy is designed to obtain more reliable results. 
Extensive experiments demonstrate that our method performs better not only against various attacks but also in image visual quality and file size increment.

\clearpage
\bibliographystyle{ACM-Reference-Format}
\balance
\bibliography{acmart}


\begin{thebibliography}{47}


\ifx \showCODEN    \undefined \def \showCODEN     #1{\unskip}     \fi
\ifx \showDOI      \undefined \def \showDOI       #1{#1}\fi
\ifx \showISBNx    \undefined \def \showISBNx     #1{\unskip}     \fi
\ifx \showISBNxiii \undefined \def \showISBNxiii  #1{\unskip}     \fi
\ifx \showISSN     \undefined \def \showISSN      #1{\unskip}     \fi
\ifx \showLCCN     \undefined \def \showLCCN      #1{\unskip}     \fi
\ifx \shownote     \undefined \def \shownote      #1{#1}          \fi
\ifx \showarticletitle \undefined \def \showarticletitle #1{#1}   \fi
\ifx \showURL      \undefined \def \showURL       {\relax}        \fi
\providecommand\bibfield[2]{#2}
\providecommand\bibinfo[2]{#2}
\providecommand\natexlab[1]{#1}
\providecommand\showeprint[2][]{arXiv:#2}

\bibitem[Ahmadi et~al\mbox{.}(2020)]%
        {ahmadi2020redmark}
\bibfield{author}{\bibinfo{person}{Mahdi Ahmadi}, \bibinfo{person}{Alireza
  Norouzi}, \bibinfo{person}{Nader Karimi}, \bibinfo{person}{Shadrokh Samavi},
  {and} \bibinfo{person}{Ali Emami}.} \bibinfo{year}{2020}\natexlab{}.
\newblock \showarticletitle{ReDMark: Framework for residual diffusion
  watermarking based on deep networks}.
\newblock \bibinfo{journal}{\emph{Expert Syst. Appl.}}  \bibinfo{volume}{146}
  (\bibinfo{year}{2020}), \bibinfo{pages}{113157}.
\newblock
\urldef\tempurl%
\url{https://doi.org/10.1016/j.eswa.2019.113157}
\showDOI{\tempurl}


\bibitem[Allwadhi et~al\mbox{.}(2022)]%
        {allwadhi2022comprehensive}
\bibfield{author}{\bibinfo{person}{Sachin Allwadhi}, \bibinfo{person}{Kamaldeep
  Joshi}, \bibinfo{person}{Ashok~Kumar Yadav}, \bibinfo{person}{Rainu Nandal},
  {and} \bibinfo{person}{Rishabh Jain}.} \bibinfo{year}{2022}\natexlab{}.
\newblock \showarticletitle{A comprehensive survey of state-of-art techniques
  in digital watermarking}. In \bibinfo{booktitle}{\emph{ICAC3N}}.
  \bibinfo{pages}{2362--2368}.
\newblock


\bibitem[Barni et~al\mbox{.}(1998)]%
        {barni1998adct}
\bibfield{author}{\bibinfo{person}{Mauro Barni}, \bibinfo{person}{Franco
  Bartolini}, \bibinfo{person}{Vito Cappellini}, {and}
  \bibinfo{person}{Alessandro Piva}.} \bibinfo{year}{1998}\natexlab{}.
\newblock \showarticletitle{A DCT-domain system for robust image watermarking}.
\newblock \bibinfo{journal}{\emph{Signal Process.}} \bibinfo{volume}{66},
  \bibinfo{number}{3} (\bibinfo{year}{1998}), \bibinfo{pages}{357--372}.
\newblock
\urldef\tempurl%
\url{https://doi.org/10.1016/S0165-1684(98)00015-2}
\showDOI{\tempurl}


\bibitem[Boroumand et~al\mbox{.}(2019)]%
        {Boroumand2019Deep}
\bibfield{author}{\bibinfo{person}{Mehdi Boroumand}, \bibinfo{person}{Mo Chen},
  {and} \bibinfo{person}{Jessica~J. Fridrich}.}
  \bibinfo{year}{2019}\natexlab{}.
\newblock \showarticletitle{Deep Residual Network for Steganalysis of Digital
  Images}.
\newblock \bibinfo{journal}{\emph{{IEEE} Trans. Inf. Forensics Secur.}}
  \bibinfo{volume}{14}, \bibinfo{number}{5} (\bibinfo{year}{2019}),
  \bibinfo{pages}{1181--1193}.
\newblock
\urldef\tempurl%
\url{https://doi.org/10.1109/TIFS.2018.2871749}
\showDOI{\tempurl}


\bibitem[Boussif et~al\mbox{.}(2021)]%
        {boussif2021novel}
\bibfield{author}{\bibinfo{person}{Mohamed Boussif}, \bibinfo{person}{Oussema
  Bouferas}, \bibinfo{person}{Noureddine Aloui}, {and} \bibinfo{person}{Adnen
  Cherif}.} \bibinfo{year}{2021}\natexlab{}.
\newblock \showarticletitle{A Novel Robust Blind AES/LWT+ DCT+ SVD-Based
  Crypto-Watermarking schema for DICOM Images Security}. In
  \bibinfo{booktitle}{\emph{DTS}}. IEEE, \bibinfo{pages}{1--6}.
\newblock


\bibitem[Camacho and Wang(2021)]%
        {castillo2021comprehensive}
\bibfield{author}{\bibinfo{person}{Ivan~Castillo Camacho} {and}
  \bibinfo{person}{Kai Wang}.} \bibinfo{year}{2021}\natexlab{}.
\newblock \showarticletitle{A Comprehensive Review of Deep-Learning-Based
  Methods for Image Forensics}.
\newblock \bibinfo{journal}{\emph{J. Imaging}} \bibinfo{volume}{7},
  \bibinfo{number}{4} (\bibinfo{year}{2021}), \bibinfo{pages}{69}.
\newblock
\urldef\tempurl%
\url{https://doi.org/10.3390/jimaging7040069}
\showDOI{\tempurl}


\bibitem[Deng et~al\mbox{.}(2009)]%
        {deng2009imagenet}
\bibfield{author}{\bibinfo{person}{Jia Deng}, \bibinfo{person}{Wei Dong},
  \bibinfo{person}{Richard Socher}, \bibinfo{person}{Li{-}Jia Li},
  \bibinfo{person}{Kai Li}, {and} \bibinfo{person}{Li Fei{-}Fei}.}
  \bibinfo{year}{2009}\natexlab{}.
\newblock \showarticletitle{ImageNet: {A} large-scale hierarchical image
  database}. In \bibinfo{booktitle}{\emph{CVPR}}. \bibinfo{publisher}{{IEEE}
  Computer Society}, \bibinfo{pages}{248--255}.
\newblock
\urldef\tempurl%
\url{https://doi.org/10.1109/CVPR.2009.5206848}
\showDOI{\tempurl}


\bibitem[Dumitrescu et~al\mbox{.}(2003)]%
        {Dumitrescu03detcetion}
\bibfield{author}{\bibinfo{person}{Sorina Dumitrescu}, \bibinfo{person}{Xiaolin
  Wu}, {and} \bibinfo{person}{Zhe Wang}.} \bibinfo{year}{2003}\natexlab{}.
\newblock \showarticletitle{Detection of {LSB} steganography via sample pair
  analysis}.
\newblock \bibinfo{journal}{\emph{{IEEE} Trans. Signal Process.}}
  \bibinfo{volume}{51}, \bibinfo{number}{7} (\bibinfo{year}{2003}),
  \bibinfo{pages}{1995--2007}.
\newblock
\urldef\tempurl%
\url{https://doi.org/10.1109/TSP.2003.812753}
\showDOI{\tempurl}


\bibitem[Fares et~al\mbox{.}(2020)]%
        {fares2020robust}
\bibfield{author}{\bibinfo{person}{Kahlessenane Fares}, \bibinfo{person}{Khaldi
  Amine}, {and} \bibinfo{person}{Euschi Salah}.}
  \bibinfo{year}{2020}\natexlab{}.
\newblock \showarticletitle{A robust blind color image watermarking based on
  Fourier transform domain}.
\newblock \bibinfo{journal}{\emph{Optik}}  \bibinfo{volume}{208}
  (\bibinfo{year}{2020}), \bibinfo{pages}{164562}.
\newblock


\bibitem[Fernandez et~al\mbox{.}(2022)]%
        {fernandez2022watermarking}
\bibfield{author}{\bibinfo{person}{Pierre Fernandez},
  \bibinfo{person}{Alexandre Sablayrolles}, \bibinfo{person}{Teddy Furon},
  \bibinfo{person}{Herv{\'{e}} J{\'{e}}gou}, {and} \bibinfo{person}{Matthijs
  Douze}.} \bibinfo{year}{2022}\natexlab{}.
\newblock \showarticletitle{Watermarking Images in Self-Supervised Latent
  Spaces}. In \bibinfo{booktitle}{\emph{ICASSP}}. \bibinfo{publisher}{{IEEE}},
  \bibinfo{pages}{3054--3058}.
\newblock
\urldef\tempurl%
\url{https://doi.org/10.1109/ICASSP43922.2022.9746058}
\showDOI{\tempurl}


\bibitem[Fridrich and Goljan(2002)]%
        {Fridrich02Practical}
\bibfield{author}{\bibinfo{person}{Jessica~J. Fridrich} {and}
  \bibinfo{person}{Miroslav Goljan}.} \bibinfo{year}{2002}\natexlab{}.
\newblock \showarticletitle{Practical steganalysis of digital images: state of
  the art}. In \bibinfo{booktitle}{\emph{SPIE}},
  \bibfield{editor}{\bibinfo{person}{Edward J.~Delp III} {and}
  \bibinfo{person}{Ping~Wah Wong}} (Eds.), Vol.~\bibinfo{volume}{4675}.
  \bibinfo{pages}{1--13}.
\newblock
\urldef\tempurl%
\url{https://doi.org/10.1117/12.465263}
\showDOI{\tempurl}


\bibitem[Fridrich et~al\mbox{.}(2001)]%
        {Fridrich01Detecting}
\bibfield{author}{\bibinfo{person}{Jessica~J. Fridrich},
  \bibinfo{person}{Miroslav Goljan}, {and} \bibinfo{person}{Rui Du}.}
  \bibinfo{year}{2001}\natexlab{}.
\newblock \showarticletitle{Detecting {LSB} Steganography in Color and
  Gray-Scale Images}.
\newblock \bibinfo{journal}{\emph{{IEEE} Multim.}} \bibinfo{volume}{8},
  \bibinfo{number}{4} (\bibinfo{year}{2001}), \bibinfo{pages}{22--28}.
\newblock
\urldef\tempurl%
\url{https://doi.org/10.1109/93.959097}
\showDOI{\tempurl}


\bibitem[Hosam(2019)]%
        {osama2019attacking}
\bibfield{author}{\bibinfo{person}{Osama Hosam}.}
  \bibinfo{year}{2019}\natexlab{}.
\newblock \showarticletitle{Attacking Image Watermarking and Steganography-A
  Survey}.
\newblock \bibinfo{journal}{\emph{International Journal of Information
  Technology and Computer Science}} \bibinfo{volume}{11}, \bibinfo{number}{3}
  (\bibinfo{year}{2019}), \bibinfo{pages}{23--37}.
\newblock


\bibitem[Hu et~al\mbox{.}(2018)]%
        {hu2020squeeze}
\bibfield{author}{\bibinfo{person}{Jie Hu}, \bibinfo{person}{Li Shen}, {and}
  \bibinfo{person}{Gang Sun}.} \bibinfo{year}{2018}\natexlab{}.
\newblock \showarticletitle{Squeeze-and-Excitation Networks}. In
  \bibinfo{booktitle}{\emph{CVPR}}. \bibinfo{publisher}{Computer Vision
  Foundation / {IEEE} Computer Society}, \bibinfo{pages}{7132--7141}.
\newblock
\urldef\tempurl%
\url{https://doi.org/10.1109/CVPR.2018.00745}
\showDOI{\tempurl}


\bibitem[Huang et~al\mbox{.}(2001)]%
        {daren2001dwt}
\bibfield{author}{\bibinfo{person}{Daren Huang}, \bibinfo{person}{Jiufen Liu},
  \bibinfo{person}{Jiwu Huang}, {and} \bibinfo{person}{Hongmei Liu}.}
  \bibinfo{year}{2001}\natexlab{}.
\newblock \showarticletitle{A Dwt-Based Image Watermarking Algorithm}. In
  \bibinfo{booktitle}{\emph{ICME}}.
\newblock
\urldef\tempurl%
\url{https://doi.org/10.1109/ICME.2001.1237719}
\showDOI{\tempurl}


\bibitem[Isola et~al\mbox{.}(2017)]%
        {phillip2017image}
\bibfield{author}{\bibinfo{person}{Phillip Isola}, \bibinfo{person}{Jun{-}Yan
  Zhu}, \bibinfo{person}{Tinghui Zhou}, {and} \bibinfo{person}{Alexei~A.
  Efros}.} \bibinfo{year}{2017}\natexlab{}.
\newblock \showarticletitle{Image-to-Image Translation with Conditional
  Adversarial Networks}. In \bibinfo{booktitle}{\emph{CVPR}}.
  \bibinfo{pages}{5967--5976}.
\newblock
\urldef\tempurl%
\url{https://doi.org/10.1109/CVPR.2017.632}
\showDOI{\tempurl}


\bibitem[Jia et~al\mbox{.}(2021)]%
        {jia2021mbrs}
\bibfield{author}{\bibinfo{person}{Zhaoyang Jia}, \bibinfo{person}{Han Fang},
  {and} \bibinfo{person}{Weiming Zhang}.} \bibinfo{year}{2021}\natexlab{}.
\newblock \showarticletitle{{MBRS:} Enhancing Robustness of DNN-based
  Watermarking by Mini-Batch of Real and Simulated {JPEG} Compression}. In
  \bibinfo{booktitle}{\emph{ACM MM}}. \bibinfo{pages}{41--49}.
\newblock
\urldef\tempurl%
\url{https://doi.org/10.1145/3474085.3475324}
\showDOI{\tempurl}


\bibitem[Kang et~al\mbox{.}(2003)]%
        {kang2003dwt}
\bibfield{author}{\bibinfo{person}{Xiangui Kang}, \bibinfo{person}{Jiwu Huang},
  \bibinfo{person}{Yun~Q. Shi}, {and} \bibinfo{person}{Yan Lin}.}
  \bibinfo{year}{2003}\natexlab{}.
\newblock \showarticletitle{A {DWT-DFT} composite watermarking scheme robust to
  both affine transform and {JPEG} compression}.
\newblock \bibinfo{journal}{\emph{{IEEE} Trans. Circuits Syst. Video Technol.}}
  \bibinfo{volume}{13}, \bibinfo{number}{8} (\bibinfo{year}{2003}),
  \bibinfo{pages}{776--786}.
\newblock
\urldef\tempurl%
\url{https://doi.org/10.1109/TCSVT.2003.815957}
\showDOI{\tempurl}


\bibitem[Ko et~al\mbox{.}(2020)]%
        {ko2020robust}
\bibfield{author}{\bibinfo{person}{Hung{-}Jui Ko}, \bibinfo{person}{Cheng{-}Ta
  Huang}, \bibinfo{person}{Gwoboa Horng}, {and} \bibinfo{person}{Shiuh{-}Jeng
  Wang}.} \bibinfo{year}{2020}\natexlab{}.
\newblock \showarticletitle{Robust and blind image watermarking in {DCT} domain
  using inter-block coefficient correlation}.
\newblock \bibinfo{journal}{\emph{Inf. Sci.}}  \bibinfo{volume}{517}
  (\bibinfo{year}{2020}), \bibinfo{pages}{128--147}.
\newblock
\urldef\tempurl%
\url{https://doi.org/10.1016/j.ins.2019.11.005}
\showDOI{\tempurl}


\bibitem[Kuznetsova et~al\mbox{.}(2018)]%
        {alina2018open}
\bibfield{author}{\bibinfo{person}{Alina Kuznetsova}, \bibinfo{person}{Hassan
  Rom}, \bibinfo{person}{Neil Alldrin}, \bibinfo{person}{Jasper R.~R.
  Uijlings}, \bibinfo{person}{Ivan Krasin}, \bibinfo{person}{Jordi
  Pont{-}Tuset}, \bibinfo{person}{Shahab Kamali}, \bibinfo{person}{Stefan
  Popov}, \bibinfo{person}{Matteo Malloci}, \bibinfo{person}{Tom Duerig}, {and}
  \bibinfo{person}{Vittorio Ferrari}.} \bibinfo{year}{2018}\natexlab{}.
\newblock \showarticletitle{The Open Images Dataset {V4:} Unified image
  classification, object detection, and visual relationship detection at
  scale}.
\newblock \bibinfo{journal}{\emph{CoRR}}  \bibinfo{volume}{abs/1811.00982}
  (\bibinfo{year}{2018}).
\newblock


\bibitem[Lin et~al\mbox{.}(2001)]%
        {lin2001rotation}
\bibfield{author}{\bibinfo{person}{Ching{-}Yung Lin}, \bibinfo{person}{Min Wu},
  \bibinfo{person}{Jeffrey~A. Bloom}, \bibinfo{person}{Ingemar~J. Cox},
  \bibinfo{person}{Matthew~L. Miller}, {and} \bibinfo{person}{Yui~Man Lui}.}
  \bibinfo{year}{2001}\natexlab{}.
\newblock \showarticletitle{Rotation, scale, and translation resilient
  watermarking for images}.
\newblock \bibinfo{journal}{\emph{{IEEE} Trans. Image Process.}}
  \bibinfo{volume}{10}, \bibinfo{number}{5} (\bibinfo{year}{2001}),
  \bibinfo{pages}{767--782}.
\newblock
\urldef\tempurl%
\url{https://doi.org/10.1109/83.918569}
\showDOI{\tempurl}


\bibitem[Lin et~al\mbox{.}(2014)]%
        {lin2014coco}
\bibfield{author}{\bibinfo{person}{Tsung{-}Yi Lin}, \bibinfo{person}{Michael
  Maire}, \bibinfo{person}{Serge~J. Belongie}, \bibinfo{person}{James Hays},
  \bibinfo{person}{Pietro Perona}, \bibinfo{person}{Deva Ramanan},
  \bibinfo{person}{Piotr Doll{\'{a}}r}, {and} \bibinfo{person}{C.~Lawrence
  Zitnick}.} \bibinfo{year}{2014}\natexlab{}.
\newblock \showarticletitle{Microsoft {COCO:} Common Objects in Context}. In
  \bibinfo{booktitle}{\emph{ECCV}}, \bibfield{editor}{\bibinfo{person}{David~J.
  Fleet}, \bibinfo{person}{Tom{\'{a}}s Pajdla}, \bibinfo{person}{Bernt
  Schiele}, {and} \bibinfo{person}{Tinne Tuytelaars}} (Eds.),
  Vol.~\bibinfo{volume}{8693}. \bibinfo{pages}{740--755}.
\newblock
\urldef\tempurl%
\url{https://doi.org/10.1007/978-3-319-10602-1\_48}
\showDOI{\tempurl}


\bibitem[Liu et~al\mbox{.}(2019)]%
        {liu2019novel}
\bibfield{author}{\bibinfo{person}{Yang Liu}, \bibinfo{person}{Mengxi Guo},
  \bibinfo{person}{Jian Zhang}, \bibinfo{person}{Yuesheng Zhu}, {and}
  \bibinfo{person}{Xiaodong Xie}.} \bibinfo{year}{2019}\natexlab{}.
\newblock \showarticletitle{A Novel Two-stage Separable Deep Learning Framework
  for Practical Blind Watermarking}. In \bibinfo{booktitle}{\emph{ACM MM}},
  \bibfield{editor}{\bibinfo{person}{Laurent Amsaleg}, \bibinfo{person}{Benoit
  Huet}, \bibinfo{person}{Martha~A. Larson}, \bibinfo{person}{Guillaume
  Gravier}, \bibinfo{person}{Hayley Hung}, \bibinfo{person}{Chong{-}Wah Ngo},
  {and} \bibinfo{person}{Wei~Tsang Ooi}} (Eds.). \bibinfo{pages}{1509--1517}.
\newblock
\urldef\tempurl%
\url{https://doi.org/10.1145/3343031.3351025}
\showDOI{\tempurl}


\bibitem[Long et~al\mbox{.}(2022)]%
        {long22frequency}
\bibfield{author}{\bibinfo{person}{Yuyang Long}, \bibinfo{person}{Qilong
  Zhang}, \bibinfo{person}{Boheng Zeng}, \bibinfo{person}{Lianli Gao},
  \bibinfo{person}{Xianglong Liu}, \bibinfo{person}{Jian Zhang}, {and}
  \bibinfo{person}{Jingkuan Song}.} \bibinfo{year}{2022}\natexlab{}.
\newblock \showarticletitle{Frequency Domain Model Augmentation for Adversarial
  Attack}. In \bibinfo{booktitle}{\emph{ECCV}}, Vol.~\bibinfo{volume}{13664}.
  \bibinfo{pages}{549--566}.
\newblock
\urldef\tempurl%
\url{https://doi.org/10.1007/978-3-031-19772-7\_32}
\showDOI{\tempurl}


\bibitem[Loshchilov and Hutter(2019)]%
        {adamw}
\bibfield{author}{\bibinfo{person}{Ilya Loshchilov} {and}
  \bibinfo{person}{Frank Hutter}.} \bibinfo{year}{2019}\natexlab{}.
\newblock \showarticletitle{Decoupled Weight Decay Regularization}. In
  \bibinfo{booktitle}{\emph{ICLR}}.
\newblock


\bibitem[Luo et~al\mbox{.}(2022)]%
        {luo2022leca}
\bibfield{author}{\bibinfo{person}{Xiyang Luo}, \bibinfo{person}{Michael
  Goebel}, \bibinfo{person}{Elnaz Barshan}, {and} \bibinfo{person}{Feng Yang}.}
  \bibinfo{year}{2022}\natexlab{}.
\newblock \showarticletitle{{LECA:} {A} Learned Approach for Efficient
  Cover-agnostic Watermarking}.
\newblock \bibinfo{journal}{\emph{CoRR}}  \bibinfo{volume}{abs/2206.10813}
  (\bibinfo{year}{2022}).
\newblock
\urldef\tempurl%
\url{https://doi.org/10.48550/arXiv.2206.10813}
\showDOI{\tempurl}


\bibitem[Mahto and Singh(2021)]%
        {mahto2021survey}
\bibfield{author}{\bibinfo{person}{Dhiran~Kumar Mahto} {and}
  \bibinfo{person}{Amit~Kumar Singh}.} \bibinfo{year}{2021}\natexlab{}.
\newblock \showarticletitle{A survey of color image watermarking:
  State-of-the-art and research directions}.
\newblock \bibinfo{journal}{\emph{Comput. Electr. Eng.}}  \bibinfo{volume}{93}
  (\bibinfo{year}{2021}), \bibinfo{pages}{107255}.
\newblock
\urldef\tempurl%
\url{https://doi.org/10.1016/j.compeleceng.2021.107255}
\showDOI{\tempurl}


\bibitem[Paszke et~al\mbox{.}(2019)]%
        {paszke2019pytorch}
\bibfield{author}{\bibinfo{person}{Adam Paszke}, \bibinfo{person}{Sam Gross},
  \bibinfo{person}{Francisco Massa}, \bibinfo{person}{Adam Lerer},
  \bibinfo{person}{James Bradbury}, \bibinfo{person}{Gregory Chanan},
  \bibinfo{person}{Trevor Killeen}, \bibinfo{person}{Zeming Lin},
  \bibinfo{person}{Natalia Gimelshein}, \bibinfo{person}{Luca Antiga},
  \bibinfo{person}{Alban Desmaison}, \bibinfo{person}{Andreas K{\"{o}}pf},
  \bibinfo{person}{Edward~Z. Yang}, \bibinfo{person}{Zachary DeVito},
  \bibinfo{person}{Martin Raison}, \bibinfo{person}{Alykhan Tejani},
  \bibinfo{person}{Sasank Chilamkurthy}, \bibinfo{person}{Benoit Steiner},
  \bibinfo{person}{Lu Fang}, \bibinfo{person}{Junjie Bai}, {and}
  \bibinfo{person}{Soumith Chintala}.} \bibinfo{year}{2019}\natexlab{}.
\newblock \showarticletitle{PyTorch: An Imperative Style, High-Performance Deep
  Learning Library}. In \bibinfo{booktitle}{\emph{NeurIPS}},
  \bibfield{editor}{\bibinfo{person}{Hanna~M. Wallach}, \bibinfo{person}{Hugo
  Larochelle}, \bibinfo{person}{Alina Beygelzimer}, \bibinfo{person}{Florence
  d'Alch{\'{e}}{-}Buc}, \bibinfo{person}{Emily~B. Fox}, {and}
  \bibinfo{person}{Roman Garnett}} (Eds.). \bibinfo{pages}{8024--8035}.
\newblock


\bibitem[Pereira and Pun(2000)]%
        {pereira2000robust}
\bibfield{author}{\bibinfo{person}{Shelby Pereira} {and}
  \bibinfo{person}{Thierry Pun}.} \bibinfo{year}{2000}\natexlab{}.
\newblock \showarticletitle{Robust template matching for affine resistant image
  watermarks}.
\newblock \bibinfo{journal}{\emph{{IEEE} Trans. Image Process.}}
  \bibinfo{volume}{9}, \bibinfo{number}{6} (\bibinfo{year}{2000}),
  \bibinfo{pages}{1123--1129}.
\newblock
\urldef\tempurl%
\url{https://doi.org/10.1109/83.846253}
\showDOI{\tempurl}


\bibitem[Peterson and Brown(1961)]%
        {peterson1961cyclic}
\bibfield{author}{\bibinfo{person}{W.~W. Peterson} {and} \bibinfo{person}{D.~T.
  Brown}.} \bibinfo{year}{1961}\natexlab{}.
\newblock \showarticletitle{Cyclic Codes for Error Detection}.
\newblock \bibinfo{journal}{\emph{IRE}} \bibinfo{volume}{49},
  \bibinfo{number}{1} (\bibinfo{year}{1961}), \bibinfo{pages}{228--235}.
\newblock


\bibitem[Qin et~al\mbox{.}(2020)]%
        {qin2020u2net}
\bibfield{author}{\bibinfo{person}{Xuebin Qin}, \bibinfo{person}{Zichen Zhang},
  \bibinfo{person}{Chenyang Huang}, \bibinfo{person}{Masood Dehghan},
  \bibinfo{person}{Osmar~R. Za{\"{\i}}ane}, {and} \bibinfo{person}{Martin
  J{\"{a}}gersand}.} \bibinfo{year}{2020}\natexlab{}.
\newblock \showarticletitle{U\({}^{\mbox{2}}\)-Net: Going deeper with nested
  U-structure for salient object detection}.
\newblock \bibinfo{journal}{\emph{Pattern Recognit.}}  \bibinfo{volume}{106}
  (\bibinfo{year}{2020}), \bibinfo{pages}{107404}.
\newblock
\urldef\tempurl%
\url{https://doi.org/10.1016/j.patcog.2020.107404}
\showDOI{\tempurl}


\bibitem[Qin et~al\mbox{.}(2019)]%
        {qin2019basnet}
\bibfield{author}{\bibinfo{person}{Xuebin Qin}, \bibinfo{person}{Zichen Zhang},
  \bibinfo{person}{Chenyang Huang}, \bibinfo{person}{Chao Gao},
  \bibinfo{person}{Masood Dehghan}, {and} \bibinfo{person}{Martin
  J{\"{a}}gersand}.} \bibinfo{year}{2019}\natexlab{}.
\newblock \showarticletitle{BASNet: Boundary-Aware Salient Object Detection}.
  In \bibinfo{booktitle}{\emph{CVPR}}. \bibinfo{pages}{7479--7489}.
\newblock
\urldef\tempurl%
\url{https://doi.org/10.1109/CVPR.2019.00766}
\showDOI{\tempurl}


\bibitem[Russell et~al\mbox{.}(2008)]%
        {bryan2007labelme}
\bibfield{author}{\bibinfo{person}{Bryan~C. Russell}, \bibinfo{person}{Antonio
  Torralba}, \bibinfo{person}{Kevin~P. Murphy}, {and}
  \bibinfo{person}{William~T. Freeman}.} \bibinfo{year}{2008}\natexlab{}.
\newblock \showarticletitle{LabelMe: {A} Database and Web-Based Tool for Image
  Annotation}.
\newblock \bibinfo{journal}{\emph{Int. J. Comput. Vis.}} \bibinfo{volume}{77},
  \bibinfo{number}{1-3} (\bibinfo{year}{2008}), \bibinfo{pages}{157--173}.
\newblock
\urldef\tempurl%
\url{https://doi.org/10.1007/s11263-007-0090-8}
\showDOI{\tempurl}


\bibitem[Shin and Song(2017)]%
        {shin2017jpeg}
\bibfield{author}{\bibinfo{person}{Richard Shin} {and} \bibinfo{person}{Dawn
  Song}.} \bibinfo{year}{2017}\natexlab{}.
\newblock \showarticletitle{Jpeg-resistant adversarial images}. In
  \bibinfo{booktitle}{\emph{NeurIPS Workshop}}.
\newblock


\bibitem[Su et~al\mbox{.}(2002)]%
        {su2002Novel}
\bibfield{author}{\bibinfo{person}{Karen Su}, \bibinfo{person}{Deepa Kundur},
  {and} \bibinfo{person}{Dimitrios Hatzinakos}.}
  \bibinfo{year}{2002}\natexlab{}.
\newblock \showarticletitle{Novel approach to collusion-resistant video
  watermarking}. In \bibinfo{booktitle}{\emph{SPIE}},
  \bibfield{editor}{\bibinfo{person}{Edward J.~Delp III} {and}
  \bibinfo{person}{Ping~Wah Wong}} (Eds.), Vol.~\bibinfo{volume}{4675}.
  \bibinfo{pages}{491--502}.
\newblock
\urldef\tempurl%
\url{https://doi.org/10.1117/12.465307}
\showDOI{\tempurl}


\bibitem[Sunesh and Kishore(2020)]%
        {kishore2020novel}
\bibfield{author}{\bibinfo{person}{Sunesh} {and} \bibinfo{person}{R.~Rama
  Kishore}.} \bibinfo{year}{2020}\natexlab{}.
\newblock \showarticletitle{A novel and efficient blind image watermarking in
  transform domain}.
\newblock \bibinfo{journal}{\emph{Procedia Computer Science}}
  \bibinfo{volume}{167} (\bibinfo{year}{2020}), \bibinfo{pages}{1505--1514}.
\newblock


\bibitem[Tan et~al\mbox{.}(2021)]%
        {Tan2021CALPA}
\bibfield{author}{\bibinfo{person}{Shunquan Tan}, \bibinfo{person}{Weilong Wu},
  \bibinfo{person}{Zilong Shao}, \bibinfo{person}{Qiushi Li},
  \bibinfo{person}{Bin Li}, {and} \bibinfo{person}{Jiwu Huang}.}
  \bibinfo{year}{2021}\natexlab{}.
\newblock \showarticletitle{{CALPA-NET:} Channel-Pruning-Assisted Deep Residual
  Network for Steganalysis of Digital Images}.
\newblock \bibinfo{journal}{\emph{{IEEE} Trans. Inf. Forensics Secur.}}
  \bibinfo{volume}{16} (\bibinfo{year}{2021}), \bibinfo{pages}{131--146}.
\newblock
\urldef\tempurl%
\url{https://doi.org/10.1109/TIFS.2020.3005304}
\showDOI{\tempurl}


\bibitem[Tancik et~al\mbox{.}(2020)]%
        {tancik2020stegastamp}
\bibfield{author}{\bibinfo{person}{Matthew Tancik}, \bibinfo{person}{Ben
  Mildenhall}, {and} \bibinfo{person}{Ren Ng}.}
  \bibinfo{year}{2020}\natexlab{}.
\newblock \showarticletitle{StegaStamp: Invisible Hyperlinks in Physical
  Photographs}. In \bibinfo{booktitle}{\emph{CVPR}}.
\newblock
\urldef\tempurl%
\url{https://doi.org/10.1109/CVPR42600.2020.00219}
\showDOI{\tempurl}


\bibitem[van Schyndel et~al\mbox{.}(1994)]%
        {van1994digital}
\bibfield{author}{\bibinfo{person}{Ron~G. van Schyndel},
  \bibinfo{person}{Andrew~Z. Tirkel}, {and} \bibinfo{person}{Charles~F.
  Osborne}.} \bibinfo{year}{1994}\natexlab{}.
\newblock \showarticletitle{A Digital Watermark}. In
  \bibinfo{booktitle}{\emph{ICIP}}. \bibinfo{pages}{86--90}.
\newblock
\urldef\tempurl%
\url{https://doi.org/10.1109/ICIP.1994.413536}
\showDOI{\tempurl}


\bibitem[Wan et~al\mbox{.}(2022)]%
        {wan2022comprehensive}
\bibfield{author}{\bibinfo{person}{Wenbo Wan}, \bibinfo{person}{Jun Wang},
  \bibinfo{person}{Yunming Zhang}, \bibinfo{person}{Jing Li},
  \bibinfo{person}{Hui Yu}, {and} \bibinfo{person}{Jiande Sun}.}
  \bibinfo{year}{2022}\natexlab{}.
\newblock \showarticletitle{A comprehensive survey on robust image
  watermarking}.
\newblock \bibinfo{journal}{\emph{Neurocomputing}}  \bibinfo{volume}{488}
  (\bibinfo{year}{2022}), \bibinfo{pages}{226--247}.
\newblock
\urldef\tempurl%
\url{https://doi.org/10.1016/j.neucom.2022.02.083}
\showDOI{\tempurl}


\bibitem[Wang et~al\mbox{.}(2021b)]%
        {wang2021admix}
\bibfield{author}{\bibinfo{person}{Xiaosen Wang}, \bibinfo{person}{Xuanran He},
  \bibinfo{person}{Jingdong Wang}, {and} \bibinfo{person}{Kun He}.}
  \bibinfo{year}{2021}\natexlab{b}.
\newblock \showarticletitle{Admix: Enhancing the Transferability of Adversarial
  Attacks}. In \bibinfo{booktitle}{\emph{ICCV}}. \bibinfo{pages}{16138--16147}.
\newblock
\urldef\tempurl%
\url{https://doi.org/10.1109/ICCV48922.2021.01585}
\showDOI{\tempurl}


\bibitem[Wang et~al\mbox{.}(2022)]%
        {wang2022triangle}
\bibfield{author}{\bibinfo{person}{Xiaosen Wang}, \bibinfo{person}{Zeliang
  Zhang}, \bibinfo{person}{Kangheng Tong}, \bibinfo{person}{Dihong Gong},
  \bibinfo{person}{Kun He}, \bibinfo{person}{Zhifeng Li}, {and}
  \bibinfo{person}{Wei Liu}.} \bibinfo{year}{2022}\natexlab{}.
\newblock \showarticletitle{Triangle Attack: {A} Query-Efficient Decision-Based
  Adversarial Attack}. In \bibinfo{booktitle}{\emph{ECCV}},
  Vol.~\bibinfo{volume}{13665}. \bibinfo{pages}{156--174}.
\newblock
\urldef\tempurl%
\url{https://doi.org/10.1007/978-3-031-20065-6\_10}
\showDOI{\tempurl}


\bibitem[Wang et~al\mbox{.}(2021a)]%
        {guo2021feature}
\bibfield{author}{\bibinfo{person}{Zhibo Wang}, \bibinfo{person}{Hengchang
  Guo}, \bibinfo{person}{Zhifei Zhang}, \bibinfo{person}{Wenxin Liu},
  \bibinfo{person}{Zhan Qin}, {and} \bibinfo{person}{Kui Ren}.}
  \bibinfo{year}{2021}\natexlab{a}.
\newblock \showarticletitle{Feature Importance-aware Transferable Adversarial
  Attacks}. In \bibinfo{booktitle}{\emph{ICCV}}. \bibinfo{pages}{7619--7628}.
\newblock
\urldef\tempurl%
\url{https://doi.org/10.1109/ICCV48922.2021.00754}
\showDOI{\tempurl}


\bibitem[Wang et~al\mbox{.}(2003)]%
        {wang2003multiscale}
\bibfield{author}{\bibinfo{person}{Zhou Wang}, \bibinfo{person}{Eero~P
  Simoncelli}, {and} \bibinfo{person}{Alan~C Bovik}.}
  \bibinfo{year}{2003}\natexlab{}.
\newblock \showarticletitle{Multiscale structural similarity for image quality
  assessment}. In \bibinfo{booktitle}{\emph{ACSSC}}, Vol.~\bibinfo{volume}{2}.
  \bibinfo{pages}{1398--1402}.
\newblock


\bibitem[Yuan et~al\mbox{.}(2022)]%
        {yuan2022natural}
\bibfield{author}{\bibinfo{person}{Shengming Yuan}, \bibinfo{person}{Qilong
  Zhang}, \bibinfo{person}{Lianli Gao}, \bibinfo{person}{Yaya Cheng}, {and}
  \bibinfo{person}{Jingkuan Song}.} \bibinfo{year}{2022}\natexlab{}.
\newblock \showarticletitle{Natural Color Fool: Towards Boosting Black-box
  Unrestricted Attacks}. In \bibinfo{booktitle}{\emph{NeurIPS}}.
\newblock


\bibitem[Zhang et~al\mbox{.}(2022)]%
        {zhang2022practical}
\bibfield{author}{\bibinfo{person}{Qilong Zhang}, \bibinfo{person}{Chaoning
  Zhang}, \bibinfo{person}{Chaoqun Li}, \bibinfo{person}{Jingkuan Song},
  \bibinfo{person}{Lianli Gao}, {and} \bibinfo{person}{Heng~Tao Shen}.}
  \bibinfo{year}{2022}\natexlab{}.
\newblock \showarticletitle{Practical No-box Adversarial Attacks with
  Training-free Hybrid Image Transformation}.
\newblock \bibinfo{journal}{\emph{CoRR}}  \bibinfo{volume}{abs/2203.04607}
  (\bibinfo{year}{2022}).
\newblock
\urldef\tempurl%
\url{https://doi.org/10.48550/arXiv.2203.04607}
\showDOI{\tempurl}


\bibitem[Zhu et~al\mbox{.}(2018)]%
        {zhu2018hidden}
\bibfield{author}{\bibinfo{person}{Jiren Zhu}, \bibinfo{person}{Russell
  Kaplan}, \bibinfo{person}{Justin Johnson}, {and} \bibinfo{person}{Li
  Fei{-}Fei}.} \bibinfo{year}{2018}\natexlab{}.
\newblock \showarticletitle{HiDDeN: Hiding Data With Deep Networks}. In
  \bibinfo{booktitle}{\emph{ECCV}} \emph{(\bibinfo{series}{Lecture Notes in
  Computer Science}, Vol.~\bibinfo{volume}{11219})},
  \bibfield{editor}{\bibinfo{person}{Vittorio Ferrari},
  \bibinfo{person}{Martial Hebert}, \bibinfo{person}{Cristian Sminchisescu},
  {and} \bibinfo{person}{Yair Weiss}} (Eds.). \bibinfo{pages}{682--697}.
\newblock
\urldef\tempurl%
\url{https://doi.org/10.1007/978-3-030-01267-0\_40}
\showDOI{\tempurl}


\end{thebibliography}
\clearpage

\appendix
\section*{Appendix}
\section{Image Resolution Distribution}
\label{appendix:a}
\begin{figure}[h]
    \centering
    \includegraphics[width=0.9\linewidth]{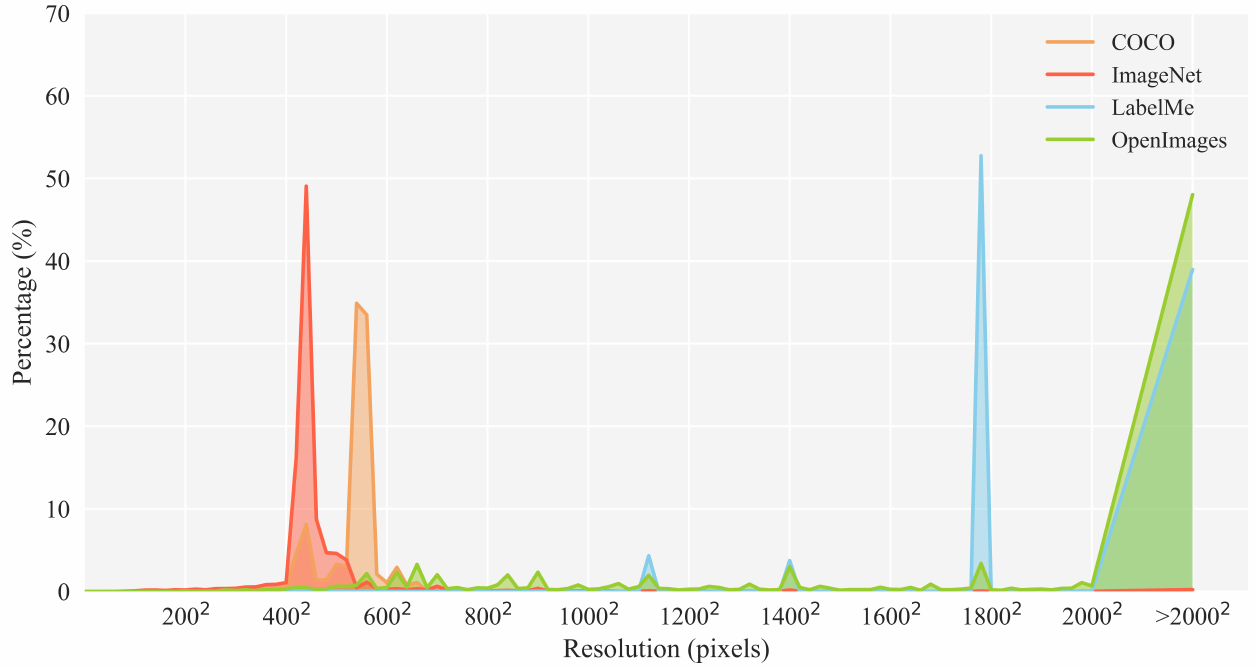}
    \caption{Resolution distribution of several common image datasets. The x-axis represents the resolution of the images, and the y-axis indicates the ratio of the number of images in the resolution range to the total dataset.}
    \label{fig:distribution}
\end{figure}

Fig.~\ref{fig:distribution} gives the resolution distribution of four image datasets used in our paper. 
Obviously, they are distributed differently.
For example, ImageNet~\cite{deng2009imagenet} is concentrated around $400\times400\sim500\times500$, while OpenImages~\cite{alina2018open} has more images with a resolution greater than $2000\times2000$. Therefore, our experiment setup---using COCO~\cite{lin2014coco} (\textit{i.e.}, concentrated around $500\times500\sim600\times600$) to train our model but evaluating our performance on other datasets---can convincingly demonstrate the generalization of our DWSF.

\begin{figure}[h]
    \centering
    \includegraphics[width=0.9\linewidth]{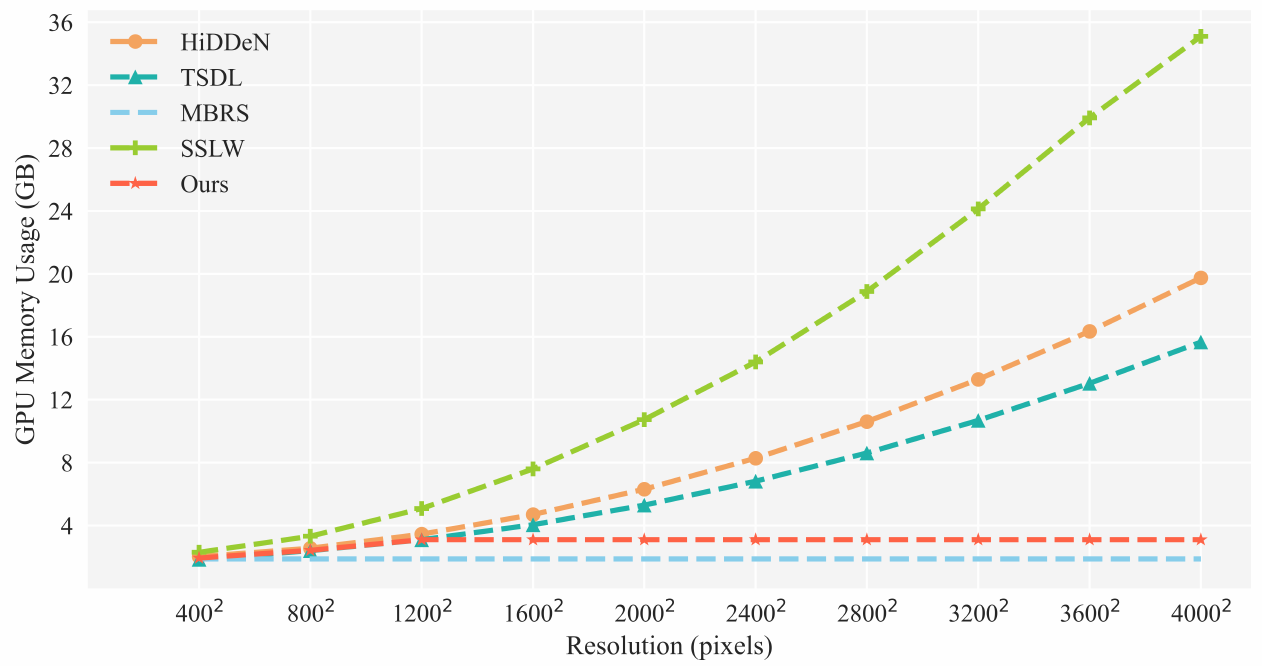}
    \caption{GPU memory used by different methods when embedding watermark messages into various resolution images.}
    \label{fig:memory}
\end{figure}
In addition, previous deep watermarking works~\cite{zhu2018hidden,liu2019novel,jia2021mbrs,fernandez2022watermarking} usually have limitations on handling various resolution images.
For example, MBRS only receives fixed-size images (\textit{e.g.}, 128$\times$128) after finish training, and we have to pre-scale the input and re-scale the output back to the original size, which brings serious distortion to the visual quality. 
Although several deep watermarking works (\textit{i.e.}, HiDDeN~\cite{zhu2018hidden}, TSDL~\cite{liu2019novel}, SSLW~\cite{fernandez2022watermarking}) can theoretically receive arbitrary resolution inputs without visual quality issue, they are still impractical due to the huge GPU memory requirements.
Fig.~\ref{fig:memory} depicts the GPU memory usage of different methods when embedding for various resolution images. Obviously, as the resolution of the input image increases, the GPU memory usage of these existing methods (\textit{i.e.}, HiDDeN, TSDL, SSLW) grow as well. For example, 
when handling an image with size 4000$\times$4000, TSDL and HiDDeN need nearly 16GB while SSLW requires more than 32GB.
By contrast, our dispersed embedding scheme can handle arbitrary resolution images with a small amount of GPU memory since we just need to handle several small-size blocks (\textit{i.e.}, maximum 20 blocks with size 128$\times$128), showing superior practicality. 

\begin{figure*}
    \centering
    \includegraphics[width=0.85\linewidth]{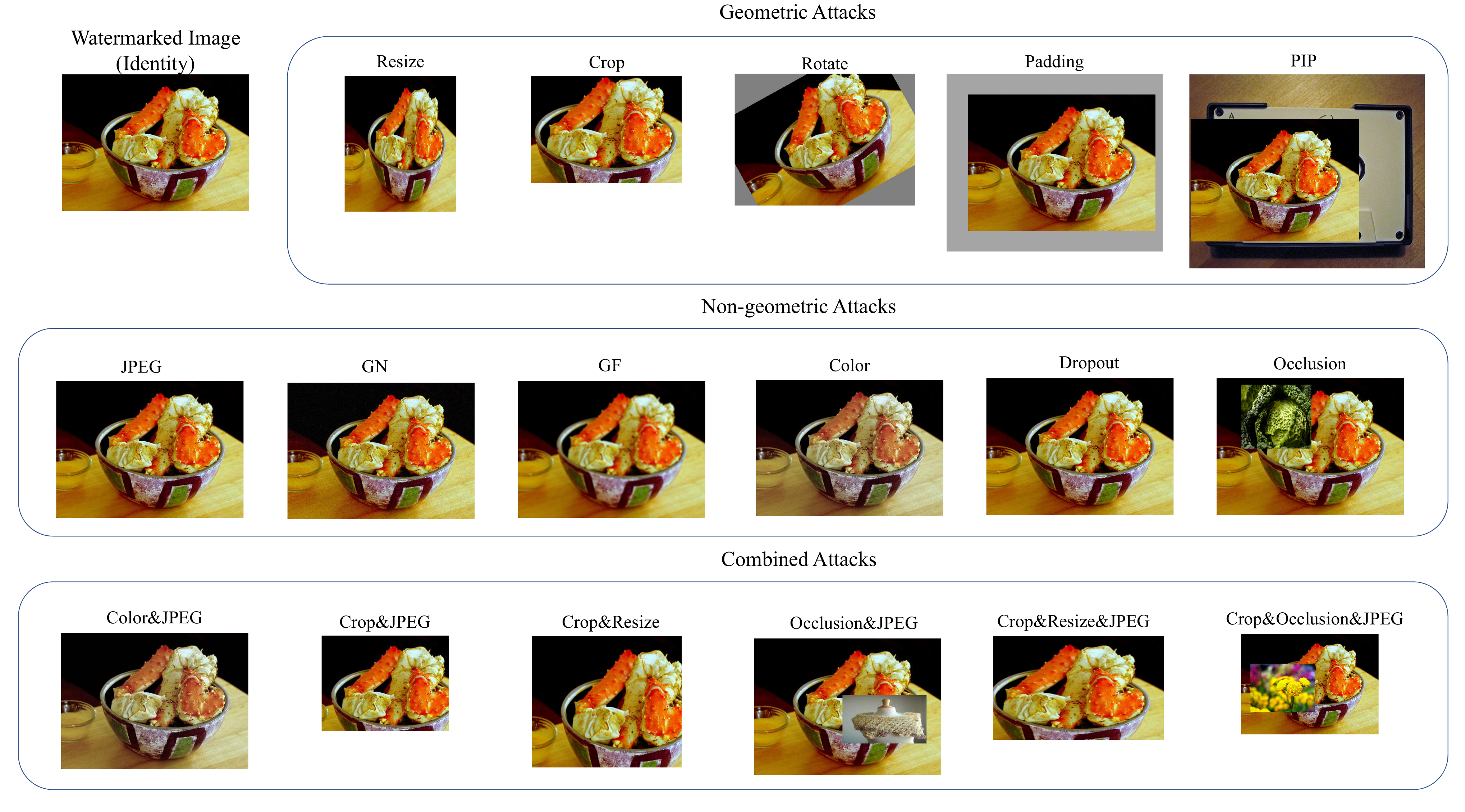}
    \caption{Visualization of geometric, non-geometric and combined attacks with random strength for watermarked image.}
    \label{fig:attack_img}
\end{figure*}

\section{Visualization of Attacked images}
\label{appendix:b}
Fig.~\ref{fig:attack_img} shows the visualization of attacks (with random strength) introduced in our paper.
Generally, geometric attacks change the size of watermarked images, while non-geometric attacks modify the pixel value of watermarked images. For combined attacks, they are more challenging for blind watermarking and are very common in the real scenario. Therefore, we suggest that the robustness of watermarking algorithms should be evaluated on sufficiently comprehensive attacks.

\begin{table}[h]
\caption{The bit check accuracy of different methods under different collusion attacks on 1,000 OpenImages watermarked images (PSNR is clipped to 35dB). Decoding is considered successful if at least one decoded message matches that of $\bm{x_1}$ or $\bm{x_2}$~/~at least one decoded message matches that of $\bm{x_1}$ and at least one decoded message matches that of $\bm{x_2}$.}
\centering
\resizebox{1\linewidth}{!}{
\begin{tabular}{cccccc}
\toprule
 & HiDDeN & TSDL & MBRS & SSLW & DWSF (Ours) \\ 
\midrule
min attack & 1.60/0.00 & 0.40/0.00 & 1.10/0.00 & 0.20/0.00 & \textbf{97.80}/\textbf{94.80} \\ 
\midrule
max attack & 1.30/0.00 & 0.30/0.00 & 0.10/0.00 & 0.20/0.00 & \textbf{97.90}/\textbf{93.60} \\ 
\midrule
mean attack & 0.00/0.00 & 0.10/0.00 & 0.60/0.00 & 0.20/0.00 & \textbf{98.50}/\textbf{94.90} \\ 
\midrule
\end{tabular}}
\label{tab:onesucc}
\end{table}
\begin{table*}
    \centering
    
    \caption{The results of different area proportion Q and the upper limit of embedded blocks (denoted by Q\%+xx) on 1,000 ImageNet watermarked images/on 1,000 OpenImages watermarked images (PSNR is clipped to 35dB).}
     \resizebox{1\linewidth}{!}{
    \begin{tabular}{ccccccccccccccc}
    \toprule
        ~ & Identity & JPEG & GN & GF & Color & Dropout & Resize & Crop & Rotate & Padding & Occlusion & PIP & AVG. & PSNR \\ \midrule
        10\%+10 & 99.98/99.99 & 96.73/99.19 & 100.0/99.99 & 99.84/99.96 & 99.80/99.92 & 99.98/99.97 & 99.47/99.92 & 93.62/99.25 & 68.97/97.25 & 99.80/99.98 & 93.44/99.06 & 99.70/99.97 & 95.94/99.54 & 45.01/47.75 \\ 
        20\%+20 & 99.98/100.0 & 97.86/99.44 & 100.0/100.0  & 99.79/99.93 & 99.85/99.97 & 99.98/99.99 & 99.87/99.88 & 97.99/99.85 & 97.47/99.33 & 100.0/100.0 & 99.01/99.85 & 99.95/100.0 & 99.31/99.85 & 42.41/46.30 \\ 
        25\%+20 & 100.0/100.0 & 98.11/99.52 & 100.0/100.0 & 99.78/99.94 & 99.95/99.95 & 99.96/100.0 & 99.80/100.0 & 98.61/99.78 & 98.20/99.64 & 99.99/100.0 & 99.04/99.91 & 100.0/100.0 & 99.45/99.90 & 42.07/46.12 \\ 
        30\%+30 & 100.0/100.0 & 98.63/99.58 & 100.0/100.0 & 99.88/100.0 & 99.93/99.93 & 100.0/100.0 & 99.97/99.97 & 99.37/99.81 & 99.01/99.72 & 100.0/100.0 & 99.67/99.92 & 100.0/99.94 & 99.71/99.91 & 40.87/45.89 \\ \bottomrule
    \end{tabular}}
    \label{tab:imagenetQ}
\end{table*}

\section{Robustness against Collusion Attacks}
\label{appendix:c}
We highlight the importance of our dispersed watermarking: can well defend against collusion attacks~\cite{su2002Novel}. In some situations, it is possible for an attacker to obtain multiple watermarked data. The attacker can often exploit this situation to remove watermarks without knowing the watermarking algorithm. This kind of attack is known as the collusion attack. In this section, we discuss three widely used collusion attacks, i.e., min attack, max attack, and mean attack. let $\bm{x_1}$ and $\bm{x_2}$ denote the watermarked image for the same raw image. The difference between $\bm{x_1}$ and $\bm{x_2}$ is the embedded message. Formally,
\begin{itemize}
    \item min attack: $\bm{\hat{x}}=\min(\bm{x_1},\bm{x_2})$ 
    \item max attack: $\bm{\hat{x}}=\max(\bm{x_1},\bm{x_2})$ 
    \item mean attack: $\bm{\hat{x}}=(\bm{x_1}+\bm{x_2})/2$ 
\end{itemize}
In Tab.~\ref{tab:onesucc}, we report our results for defending against collusion attacks. With our dispersed watermarking, we can significantly reduce the impact of the collusion attack on our watermarked image, as there is a very low probability that the embedded regions of the two images overlap. The high bit check accuracy, close to \textbf{100\%} (Ours) vs. close to 0\% (Others), convincingly demonstrates the effectiveness of our DWSF.

\section{Discussion on area proportion}
\label{appendix:d}
For dispersed embedding, in our main paper, the area
proportion Q is 25\% (with an upper limit of 20 blocks). In this section, we conduct an experiment on ImageNet and OpenImages to explain why we choose this setting. As shown in Tab.~\ref{tab:imagenetQ}, we can observe a higher $Q$ and upper limit of embedded blocks usually bring a higher bit accuracy but a lower PSNR. If proportion Q is 10\% (with an upper limit of 10 blocks), the robustness to resist the Rotate decreases, especially when the image resolution is small (\textit{e.g.}, ImageNet). If area proportion is large than 30\% (with an upper limit of 30 blocks), the performance gain is not as significant, but comes at the cost of poorer visual quality (\textit{i.e.}, lower PSNR). Therefore, we set $Q=25\%$ and the upper limit of embedded blocks to 20 so that we can outperform the powerful compared methods while maintaining high visual quality.
\begin{algorithm}[h]
    \caption{Message Fusion}
    \KwIn{The decoded results $\bm{M'} \in [0.0,1.0]^{L}$, the number of the decoded results $N$, the upper limit of the bit difference threshold $T$, and the smallest number $K$ of messages for fusion.}
    \KwOut{The final message $\bm{\Tilde{M}}  \in \{0,1\}^{L}$.}
    \For{ i = 0 to N - 1}
        {
            \For{ j = 0 to N-1}
            {

                $\mathcal{D}_{i,j}=\sum (Binary(\bm{M_i'})-Binary(\bm{M_j'}))^2$\\

            }
        }
    \For{ t = 0 to T}
    {
    $\Tilde{i_t} = \mathop{\arg\max}\limits_{0\leq i\leq N-1} |\bm{S_i} \leq t|$\\
       \If{$|\bm{S_{\Tilde{i_t}}} \leq t| \geq K$}
            {
            $\bm{\Tilde{M}}=Binary(Mean(\{\bm{M_j'}|\mathcal{D}_{\Tilde{i_t},j}\leq t\}))$\\
            \Return{$\bm{\Tilde{M}}$}
            }
    }
    $\bm{\Tilde{M}} = Binary(MEAN(\{\bm{M_j'}|0\leq j \leq N-1\}))$.\\
    \Return{$\bm{\Tilde{M}}$}.\\
\label{alg:message_fusion}
\end{algorithm}

\end{document}